\documentclass[lettersize,journal]{IEEEtran}
\usepackage{amsmath,amsfonts}
\usepackage{algorithmic}
\usepackage{algorithm}
\usepackage{array}
\usepackage[caption=false,font=normalsize,labelfont=sf,textfont=sf]{subfig}
\usepackage{textcomp}
\usepackage{stfloats}
\usepackage{url}
\usepackage{color}
\usepackage{multirow}
\usepackage{tabularray}
\usepackage{verbatim}
\usepackage{graphicx}

\usepackage{bbding}
\usepackage{booktabs}
\usepackage{cite}
\begin{document}

\title{Low-Light Object Tracking: A Benchmark}

\author{Pengzhi Zhong, Xiaoyu Guo, Defeng Huang,  Xiaojun Peng, Yian Li, Qijun Zhao, and Shuiwang Li$^*$\thanks{*Corresponding author.}

\thanks{Pengzhi Zhong, Xiaoyu Guo, Defeng Huang, Xiaojun Peng, Yian Li, Shuiwang Li are with the Guilin University of Technology, Guilin 541006,
China. (e-mail: zhongpengzhi@glut.edu.cn; guoxiaoyu@glut.edu.cn; huangdefeng@glut.edu.cn; pengxiaojun@glut.edu.cn; liyian@glut.edu.cn; 
lishuiwang0721@163.com).}
\thanks{Qijun Zhao is with College of Computer Science, Sichuan University, Sichuan 610065, China (e-mail: qjzhao@scu.edu.cn).}

}

\markboth{}
{}

\maketitle

\begin{abstract}
In recent years, the field of visual tracking has made significant progress with the application of large-scale training datasets. These datasets have supported the development of sophisticated algorithms, enhancing the accuracy and stability of visual object tracking. However, most research has primarily focused on favorable illumination circumstances, neglecting the challenges of tracking in low-ligh environments. In low-light scenes, lighting may change dramatically, targets may lack distinct texture features, and in some scenarios, targets may not be directly observable. These factors can lead to a severe decline in tracking performance. To address this issue, we introduce LLOT, a benchmark specifically designed for Low-Light Object Tracking. LLOT comprises 269 challenging sequences with a total of over 132K frames, each carefully annotated with bounding boxes. This specially designed dataset aims to promote innovation and advancement in object tracking techniques for low-light conditions, addressing challenges not adequately covered by existing benchmarks. 
To assess the performance of existing methods on LLOT, we conducted extensive tests on 39 state-of-the-art tracking algorithms. The results highlight a considerable gap in low-light tracking performance. In response, we propose H-DCPT, a novel tracker that incorporates historical and darkness clue prompts to set a stronger baseline. H-DCPT outperformed all 39 evaluated methods in our experiments, demonstrating significant improvements.
We hope that our benchmark and H-DCPT will stimulate the development of novel and accurate methods for tracking objects in low-light conditions. The LLOT and code are available
at \url{https://github.com/OpenCodeGithub/H-DCPT}.
\end{abstract}


\begin{figure*}[ht]
	\centering
	{
		\begin{minipage}[t]{1\textwidth}
			\centering
			\includegraphics[width=1\textwidth]{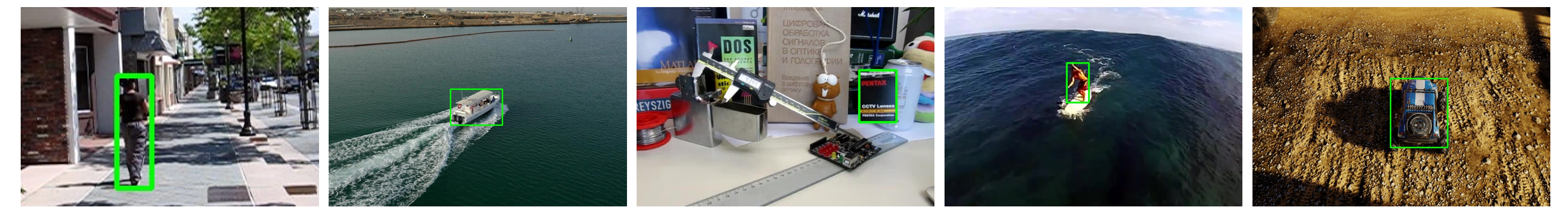}
			\centerline{\footnotesize (a) Example of object tracking under favorable illumination conditions.}
	\end{minipage}}
	{
		\begin{minipage}[t]{1\textwidth}
			\centering
			\includegraphics[width=1\textwidth]{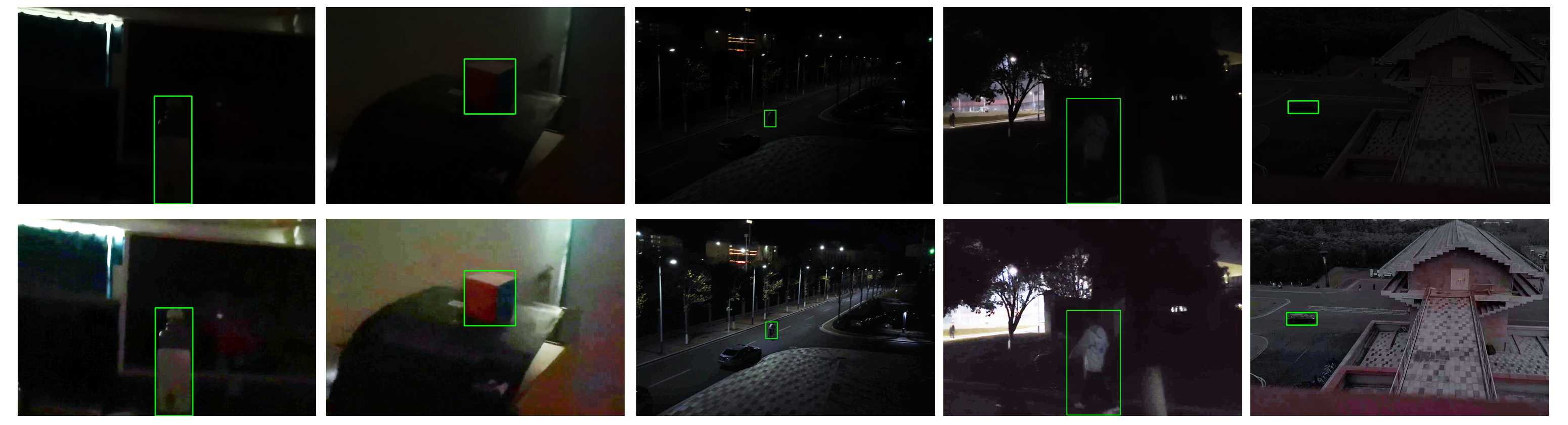}
			\centerline{\footnotesize (b) Example of object tracking under low-light conditions.}
	\end{minipage}}
	\caption{(a) Object tracking under favorable illumination conditions and (b) object tracking under low-light conditions. In Figure (b), the second row exhibits the results after image enhancement processing. Compared to well lighting conditions, low-light environments significantly increase the difficulty of object tracking. This is because low-light environments often present additional challenges such as high levels of noise, color distortions, lower contrast and visibility. }
	\label{fig:LLOT_example}
\end{figure*}

\section{Introduction}
\IEEEPARstart{V}{isual} Object Tracking (VOT) is one of the most fundamental tasks in the field of computer vision, aiming to estimate the position and scale of an initially given target in an image sequence \cite{sm2021review,zhang2020correlation,lijia2020adaptive,danelljan2020probabilistic,marvasti2021deep,gao2022aiatrack,yuan2023altrack,zhu2022siamese-orpn}. This technique is widely used in various applications such as video surveillance, autonomous driving, motion analysis, medical imaging, and drone navigation \cite{sm2021review,zhang2020correlation,lijia2020adaptive,liu2021moving,liu2021sg,tang2022ranking,danelljan2020probabilistic}. While object tracking has been extensively studied and significantly advanced over the past decades, much of the research has focused on generic objects, leaving more specialized and challenging scenes relatively underexplored \cite{zhang2022tsfmo,wu2024dtto}. One particularly challenging yet under-explored scenario is tracking objects in low-light conditions. Current datasets are primarily collected in well-lit environments, failing to adequately consider the common but more challenging low-light conditions. Consequently, models trained and tested on these datasets often exhibit significant performance degradation in low-light environments. This is due to the unique challenges posed by low-light conditions, such as high-level noise, color distortion, low contrast and visibility \cite{zhu2023dcpt,li2022uavdark135,zhang2024brve,sun2023modified,fu2023pairlie}, all of which make object identification and tracking more difficult. Fig.\ref{fig:LLOT_example} illustrates object tracking examples under favorable lighting conditions (a) and low-light environments (b). In Fig. \ref{fig:LLOT_example}(b), the first two columns depict low-light scenes indoors during the daytime, while the last three columns showcase nighttime scenarios. The bottom row demonstrates the results of image enhancement processing. It is evident from the comparison that object tracking faces greater challenges in low-light conditions compared to well lighting scenarios. Accurately tracking objects under low-light conditions, whether indoors or outdoors, is crucial for enhancing the efficiency and safety of systems in various fields such as transportation, rescue, and military applications. For instance, in the field of traffic monitoring \cite{robert2009night,robert2009video,chen2010real}, cameras may encounter issues such as increased image noise and reduced contrast under low-light conditions like nighttime or heavy rain, making it more difficult to detect and track vehicles or pedestrians. By optimizing tracking algorithms using low-light datasets, we can significantly improve the accuracy and reliability of surveillance in low-light scenarios, thereby ensuring traffic safety and order. In rescue operations \cite{chatziparaschis2020aerial,tang2024review,queralta2020collaborative}, unmanned aerial vehicles (UAVs) or robots can replace humans in navigating and performing hazardous detection tasks in low-light environments such as in mines, caves and other low-light environments. In these settings, lighting is extremely limited, and traditional daytime trackers cannot accurately navigate or identify targets. By employing low-light object tracking algorithms, UAVs and robots can recognize paths and targets in the dark, thereby improving the safety, success rate, and efficiency of rescue missions. For observing wild animals \cite{hughey2018challenges,tuia2022perspectives}, their nighttime activity is higher, but illumination conditions are poorer, limiting direct observation. Through low-light imaging devices and object tracking algorithms, it becomes possible to monitor animal behavior patterns over extended periods. For instance, tracking the nighttime hunting routes of Siberian tigers aids in studying their ecological habits; observing the movements of small nocturnal animals can help decipher their predator evasion strategies. Additionally, monitoring the nighttime activities of endangered species in protected areas requires low-light visual support. Tracking algorithms optimized based on low-light datasets can improve the recognition capabilities of field equipment in darkness, providing more effective information for conservation efforts. In military and border security domain \cite{abba2024real,ahmed2021real,dilshad2020applications,budiharto2020design}, low-light object tracking technology can significantly enhance the effectiveness of nighttime border surveillance. It aids in real-time monitoring of border activities through low-light cameras and optimized algorithms, helping to prevent illegal entry and terrorist activities.

Object tracking under low-light conditions plays an important role in many application scenarios but also poses technical challenges. Experience shows that the motion characteristics of targets vary under different illumination environments, and low-light environments more easily cause difficulties such as illumination variation and motion blur \cite{li2022uavdark135,zhang2024brve}. While most mainstream tracking algorithms today are designed based on the assumption of favorable illumination conditions, their performance in low-light scenes is not satisfactory, making it difficult to meet the needs of these applications. Therefore, it is necessary to design specialized benchmarks and optimization algorithms based on extensive research on the distinctive characteristics of object tracking mechanisms in low-light environments. In this study, we introduce the Low-Light Object Tracking (LLOT) benchmark, aimed at advancing research in this challenging field. LLOT dataset contains 269 video sequences with over 132k image frames, covering a wide range of object categories and complex indoor and outdoor scenes. Each sequence has been manually annotated and verified, along with detailed labeling of 12 challenge attributes for algorithm performance testing and analysis. Furthermore, We conducte an a comprehensive of 39 state-of-the-art tracking algorithms using the LLOT benchmark. The results indicate that existing methods generally struggle with the complexities introduced by low-light environments, highlighting the significant potential for improvement in this field. By releasing LLOT, we aim to foster innovation in tracking technology to address the unique challenges posed by low-light objects. This effort is geared towards enhancing the accuracy and reliability of visual tracking systems. The main contributions of this paper are as follows:
\begin{itemize} 

\item We introduce LLOT (Low-Light Object Tracking), which to our knowledge is the first comprehensive benchmark specifically designed for general object tracking in low-light conditions. The LLOT dataset comprises 269 carefully annotated indoor and outdoor video sequences, encompassing 32 object categories. This comprehensive dataset aims to advance research and applications in low-light object tracking.
\item We conducte a comprehensive evaluation of 39 state-of-the-art tracking algorithms on LLOT. This evaluation highlights the limitations of current tracking algorithms, establishes new performance benchmarks, and motivates future research efforts in the field of low-light object tracking.

\item We propose a novel tracker, H-DCPT, which ingeniously combines historical and dark clue prompts to improve performance. H-DCPT outperformed existing state-of-the-art trackers on the LLOT, providing a stronger baseline for future research in low-light object tracking.
\end{itemize}

\section{Related Works}
\subsection{Low-Light Image Enhancement}

Images captured in low-light environments invariably suffer from multiple distortions, such as poor visibility, low contrast, and lack of edge texture. Consequently, low-light image enhancement receives widespread attention in recent decades. One of the commonly used image enhancement methods \cite{celik2011contextual,lee2013contrast} is to adjust the pixel intensity distribution by modifying the image histogram, thereby enhancing image contrast and the differentiation between light and dark areas, achieving the goal of enhancement. Guo et al. \cite{guo2016lime} first estimate the illumination condition of the input image based on the maximum pixel value in the R, G, and B channels. Then, they use the estimated and refined illumination map to perform illumination correction on the input image, based on the Retinex-based model \cite{land1977retinex}. Wei et al. \cite{wei2018deep} establish a real-world low-light image enhancement benchmark featuring paired normal-light and low-light images. They then train an end-to-end model using a fully supervised approach. In \cite{wang2019underexposed}, Wang et al. present an approach for network structure optimization utilizing an intermediate illumination map, thereby mapping the relationship between the low-light input and expected output through the illumination map to realize enhancement. However, these learning-based LIE methods rely on paired normal-light and low-light images. To address the issue of insufficient paired normal-light/low-light data, Jiang et al. \cite{jiang2021enlightengan} adopt an unsupervised and unpaired approach to train a Generative Adversarial Network (GAN). Recently, Fu et al. \cite{fu2023pairlie} propose an unsupervised low-light image enhancement method that learns adaptive priors from pairs of low-light images, combining Retinex theory and self-supervised mechanisms to achieve efficient image detail recovery and contrast enhancement. Although these enhancement algorithms perform well in visual effects, their effectiveness in actual low-light tracking tasks remains unsatisfactory. This is mainly because the optimization objectives of these methods do not fully align with visual tracking requirements.

\subsection{Visual Tracking Algorithms}
\textbf{Daytime Trackers.} Object tracking is a crucial task in computer vision. In recent decades, visual trackers based on Discriminative Correlation Filters (DCF) \cite{fu2021correlation,li2022learning,li2020autotrack,huang2019arcf} and Deep Learning (DL) \cite{marvasti2021deep,cao2021hift,li2023aba-vitrack,lan2023procontext,cai2024hiptrack} have developed rapidly, but primarily for daytime scenarios. DCF-based trackers, due to their difficulty in end-to-end training, often rely on limited handcrafted features, resulting in poor performance under adverse conditions. Benefiting from the emergence of numerous large datasets, DL-based trackers have gained attention for their powerful ability to automatically learn features using neural networks. SiamFC, as one of the typical representatives of DL trackers, innovatively applies a siamese network architecture to visual object tracking and achieves significant progress in this field. Studies such as \cite{li2019siamrpn++,guo2020siamcar,xu2020siamfc++,cao2021siamapn++,guo2021siamgat,cao2022tctrack} have shown improvements in tracking accuracy and robustness. Recently, Visual Transformers (ViTs) have demonstrated significant potential in simplifying and unifying general visual tracking frameworks. For instance, OStrack \cite{ye2022ostrack} integrates feature extraction with relationship modeling and incorporates an early elimination module, resulting in a state-of-the-art tracking framework. ProContEXT \cite{lan2023procontext} utilizes context-aware self-attention modules to encode spatiotemporal context, progressively updating multi-scale static and dynamic templates for precise tracking. Aba-VTrack \cite{li2023aba-vitrack} proposes a lightweight deep learning tracker based on ViT, utilizing adaptive token computation methods and background awareness to minimize inference time, achieving low-latency object tracking. The latest HIPTrack \cite{cai2024hiptrack} provides a historical prompt network, offering comprehensive and precise tracking guidance through refined historical foreground masks and target visual features.
\begin{figure}[!t]
\centering
\includegraphics[width=3.4in]{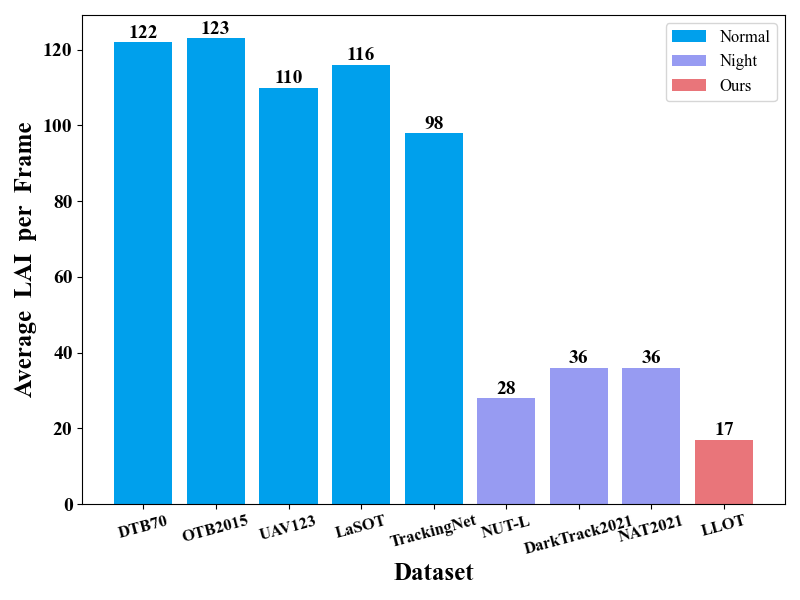}
\caption{Comparison of the average LAI (Low Ambient Illuminationg \cite{ye2022nat2021}) per frame of common datasets, nighttime datasets and our LLOT dataset.}
\label{fig:average_LAI}
\end{figure}

\textbf{Nighttime Trackers.} Compared to bright daytime environments, nighttime conditions present numerous challenges: a loss of image edges and detail, increased noise, and low contrast \cite{zhu2023dcpt,li2022uavdark135,zhang2024brve,sun2023modified,fu2023pairlie}. These characteristics of low-light objects pose major difficulties in the design and implementation of tracking algorithms. Although existing trackers perform excellently in daytime scenarios, their performance often degrades significantly or even fails in low-light nighttime environments. Consequently, nighttime tracking have garnered increasing attention. \cite{ye2021darklighter,fu2022highlightnet,ye2022sct} proposes a two-step "enhance-then-track" tracking method, which first designs a low-light image enhancement algorithm suitable for nighttime scenes to improve image quality, and then uses daytime trackers for object tracking. ADTrack \cite{li2021adtrack} integrates a low-light image enhancer into a CF-based tracker to achieve robust tracking at night. While these methods are simple and effective, they incur additional computational costs, and the separate enhancement and tracking approach is not conducive to building end-to-end trainable nighttime tracking methods. To address this issue, UDAT \cite{ye2022udat} proposes an unsupervised domain adaptation framework that achieves nighttime aerial object tracking through object discovery and transformer alignment features. However, the lack of high-quality tracking data also limits its performance. Recently, SAM-DA \cite{fu2023samda} utilizes the SAM \cite{kirillov2023sam} model to generate a large number of high-quality target domain training samples, significantly expanding domain adaptation capabilities. DCPT \cite{zhu2023dcpt} achieves robust nighttime drone tracking by learning to generate dark cue prompts and adaptively fusing these prompts with daytime trackers through a gated feature aggregation mechanism.

\subsection{Visual Tracking Benchmarks}
Benchmark datasets play a crucial role in the development of visual tracking technology, serving as key standards for evaluating the performance of tracking methods. Currently, benchmark datasets are mainly divided into two categories: general benchmarks and specific benchmarks \cite{fan2021totb}.

\textbf{General benchmarks.} General benchmark datasets are mostly utilized to assess the performance of tracker in universal scenarios. Among them, OTB is one of the early widely used object tracking benchmark datasets which adopts dense bounding box (BBox) annotations. OTB-13 \cite{wu2013otb} includes 51 video sequences, while OTB-15 \cite{wu2015otb} extends this to 100 video sequences, all labeled with 11 attributes for comprehensive evaluation of trackers. VOT \cite{kristan2016vot} is an annual competition dataset that provides over 62 sequences with more than 20K frames in 2022. The NUS-PRO \cite{li2015nuspro} dataset is mainly used to address challenges in pedestrian and rigid object tracking, including 365 challenging sequences captured by moving cameras, covering 12 different types of objects. With the widespread application of deep learning in object tracking, researchers have proposed some larger-scale datasets to meet the training needs of models. For example: LaSOT \cite{fan2021lasot}, A dataset for long-term and large-scale tracking evaluation, including 1360 long video sequences with a total of over 38.7 million frames, aimed at evaluating the performance of tracking algorithms in long-term scenarios. TrackingNet \cite{muller2018trackingnet} composed of over 30,000 video sequences, used to evaluate the performance of tracking algorithms in real-world scenarios. And GOT-10k \cite{huang2019got10k} including over 10,000 video sequences and 1.5 million manually annotated boxes. It uses the WordNet semantic system for classification guidance, covering more than 560 types of motion object classes. It aims to provide a unified training and evaluation platform for class-agnostic, generic short-term trackers.

\begin{figure*}[t]
\centering
\includegraphics[width=1\textwidth]{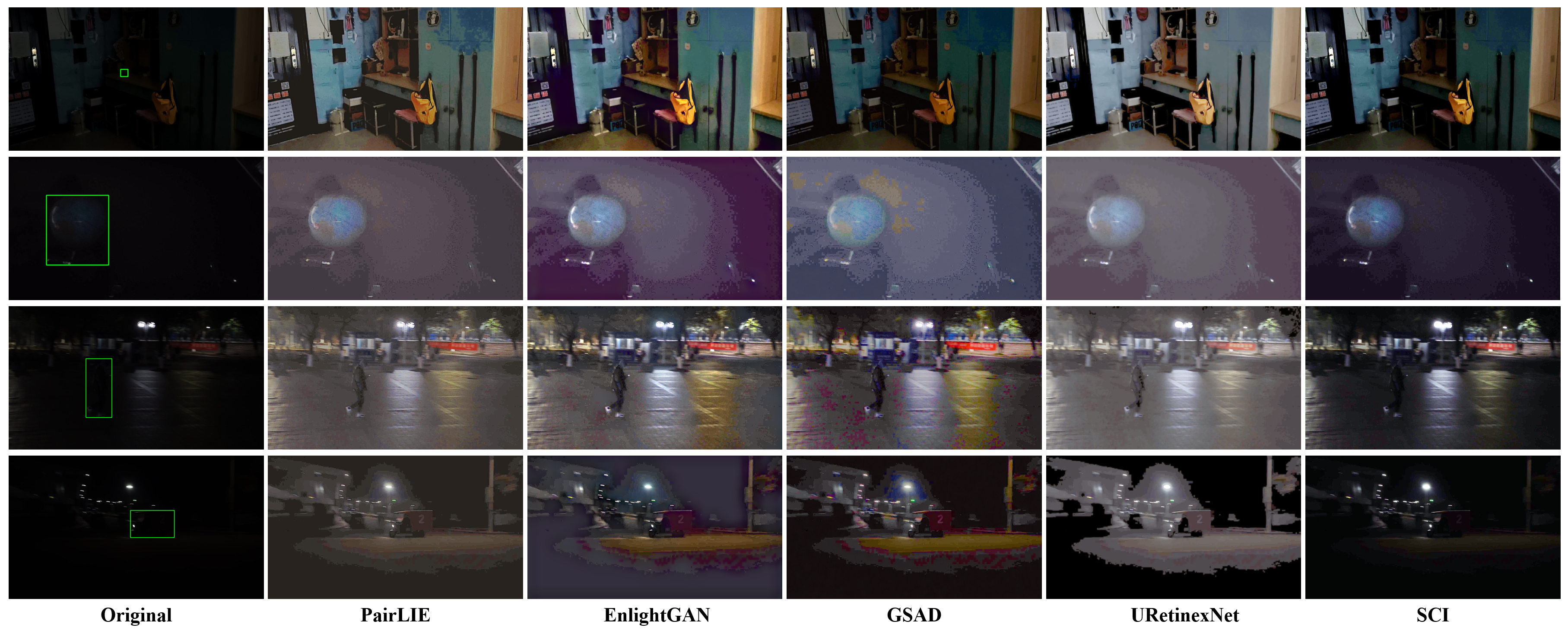}

\caption{Visual comparison of different low-light image enhancement methods on the LLOT dataset. The comparison demonstrates that the SCI method exhibits significant advantages when processing the LLOT dataset: (1) enhanced images show clearer object edge contours; (2) image noise is less pronounced compared to other methods; and (3) the enhancement effect more closely resembles natural human visual perception. These characteristics make SCI an ideal choice for LLOT preprocessing.}
\label{fig:enhanceAl}
\end{figure*}

\textbf{Specific benchmarks.} Specific benchmark datasets are designed to evaluate the tracking of particular objects. For example, UAV123 \cite{mueller2016uav123} and DTB70 \cite{li2017dtb70} contain 123 and 70 low-altitude target tracking video sequences captured by UAVs, respectively, they specifically designed to evaluate tracking algorithms in UAVs scenarios. The RGBT234 \cite{li2019rgb234} dataset provides 234 video sequences totaling approximately 233,800 frames, with each sequence containing both RGB and thermal infrared video data. TOTB \cite{fan2021totb} primarily focuses on transparent object tracking.For night-time tracking scenarios, UAVDark135 \cite{li2022uavdark135}, as the first UAV night-time tracking dataset, includes over 125,000 manually annotated images. DarkTrack2021 \cite{ye2022darktrack2021} comprises 110 challenging night-time sequences totaling over 100,000 frames. The NAT2021 \cite{ye2022nat2021} dataset aims to comprehensively evaluate night-time aerial tracking performance while providing ample unlabeled night-time aerial video data for unsupervised learning. Recently emerged TSFMO \cite{zhang2022tsfmo} focuses on tracking small and fast-moving objects, while 360VOT \cite{huang2023360vot}, as the first large-scale benchmark dataset for omnidirectional visual object tracking, includes 120 sequences with up to 113K high-resolution frames. Additionally, the LMOT \cite{wang2024lmot} dataset specifically addresses low-light multi-object tracking in outdoor environments. Existing low-light tracking datasets such as UAVDark135 \cite{li2022uavdark135}, DarkTrack2021 \cite{ye2022darktrack2021}, and NAT2021 \cite{ye2022nat2021} provide valuable resources for object tracking research under night-time conditions, they primarily concentrate on UAV-perspective scenarios. However, there remains a gap in comprehensive datasets that cover a wide range of low-light tracking scenarios beyond UAV perspectives and outdoor settings.
With the widespread application of low-light object tracking in security surveillance, autonomous driving, military and other fields, the demand for more diversified low-light tracking datasets is growing. To address this gap, we propose a new low-light object tracking dataset called LLOT (Low-Light Object Tracking). The significant difference between LLOT dataset and existing nighttime tracking datasets lies in: it uses ordinary perspectives to collect various indoor and outdoor low-light scenes, and particularly contains numerous objects that are difficult to identify with naked eyes, which is closer to actual challenging application environments. To better analyze the differences between common datasets and low-light (including nighttime) datasets, we use the LAI (Low Ambient Illuminationg \cite{ye2022nat2021}) metric for comparison. We compare the average LAI value per frame across common datasets (DTB70, OTB2015, LaSOT, TrackingNet), nighttime datasets (NUT-L, DarkTrack2021, NAT2021), and our LLOT dataset, as shown in Figure \ref{fig:average_LAI}. The figure clearly illustrates that most commom datasets have LAI values above 100, with the exception of TrackingNet, which still reaches an average LAI of 98. Nighttime datasets show significantly lower LAI values, with NUT-L, DarkTrack2021, and NAT2021 reaching a maximum of only 36. Our LLOT dataset presents an even greater challenge, with an average LAI value of just 17, substantially lower than other datasets. This comparison highlights the uniqueness and challenging nature of the LLOT dataset in low-light scenarios. By providing such a diversified and challenging dataset, we aim to promote the development of low-light object tracking algorithms and improve their performance in real world low illumination conditions.

\begin{figure*}[!t]
\centering
\includegraphics[width=1\textwidth]{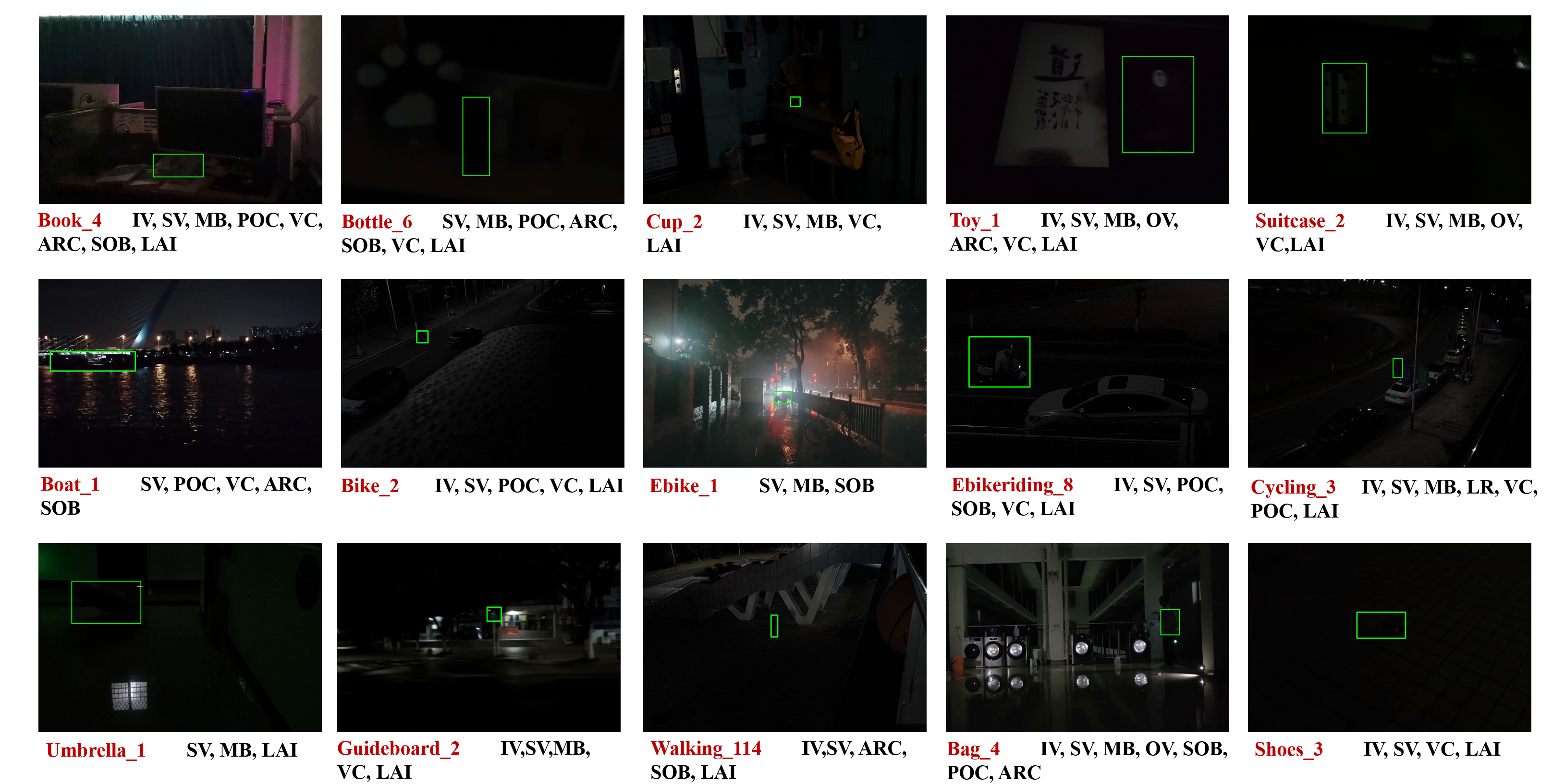}
\caption{Some examples of box annotations for LLOT.}
\label{fig:lable}
\vspace{-0.15in}
\end{figure*}

\section{A New Benchmark Dataset: LLOT}
This section provides a detailed description of the construction process of the LLOT, including methods for low-light video collection, annotation procedures, and the 12 challenge attributes. The goal is to develop a high-quality benchmark dataset for research in low-light visual object tracking.

\subsection{Collection}

Compared to well-illuminated videos, low-light videos resources suitable for object tracking are significantly scarcer on the Internet. Therefore, we only collect a small number of candidate videos online and devote considerable time and effort to shoot hundreds of candidate videos ourselves. When ranking and filtering videos suitable for low-light object tracking, we require the videos to have the following five basic characteristics: 1) Sufficient video length to evaluate the tracking performance of algorithms; 2) The background has a certain degree of change to simulate the complexity of real scenes; 3) The object has sufficient relative motion to provide diversified tracking scenarios; 4) The overall video is in a low-light environment; 5) The tracked object should not occupy an excessively large portion of the image. Building upon these fundamental criteria, we paid particular attention to video sequences that contain additional challenging factors. These challenges may include partial occlusion, motion blur, illumination changes or scale changes, etc. After collecting, processing, and annotating video data, we ultimately select 269 sequences for the LLOT benchmark. These sequences include low-light indoor and outdoor scenes, covering a variety of target categories, including human (e.g., runner, pedestrian, basketball player, football player), animals (e.g., dogs, chicken), rigid objects (e.g., bag, basin, basketball, bike, bin, boat, book, bottle, broom, car, chair, cube, cup, e-bike, globe, guideboard, hairdryer, shoe, suitcase, tourbus, toy, tricycle, truck, umbrella), and human-vehicle combinations (e.g., person \& bike, person \& e-bike). In Fig. \ref{fig:LLOT_example}, we present sample images from low-light scenes in the LLOT dataset.

\begin{table}[!t]
\centering
\caption{Attribute Description: LLOT not only incorporates the 11 standard attributes widely used in existing benchmarks but also introduces 1 additional illumination-related attribute.}
\label{tab:attr_meanings}
\setlength{\tabcolsep}{2pt} 
\begin{tabular}{@{}c!{\vrule width 1pt}p{7.0cm}@{}}
\toprule
\textbf{Attribute} & \multicolumn{1}{c}{\textbf{Meaning}} \\
\midrule
IV                   & Illumination changes in the target area.                          \\

OV                   & Target leaves the camera's field of view completely.              \\ 

LR                   & Target annotation area less than 1,000 pixels.                     \\ 

POC                  & Target is partially hidden by other objects in the scene.         \\ 

ROT                  & The target in the image experiences rotation.                     \\ 

FOC                  & Target is completely hidden by other objects in the scene.        \\ 
VC                   & Change in camera angle or perspective relative to the target.     \\ 
\multirow{2}{*}{SV}  & The ratio of the annotated target area between the first and~     \\
                     & current frames exceeds the range [0.5, 2].                        \\ 

\multirow{2}{*}{MB}  & Blurring of the target due to motion of the camera or the~        \\
                     & Target.                                                           \\ 

\multirow{2}{*}{ARC} & The change in the target's aspect ratio between the first and     \\
                     & current frames exceeds the range of [0.5, 2].                     \\

\multirow{2}{*}{SOB} & The presence of objects with a similar appearance to the     \\
                     & target.                     \\ 

\multirow{2}{*}{LAI} & Average pixel intensity in the local area centered on the object  \\
                     & below 20.                                                         \\
\bottomrule

\end{tabular}

\end{table}

\subsection{Annotation}
All frames in the LLOT are manually annotated by professionals familiar with object tracking. Given the presence of many images that are difficult to discern with the naked eye, we first apply low-light image enhancement \cite{ma2022sci} to the raw images before annotation. To obtain high-quality low-light enhanced images, we compare five state-of-the-art low-light enhancement algorithms: PairLIE \cite{fu2023pairlie}, EnlightGAN \cite{jiang2021enlightengan}, GSAD \cite{hou2024gsad}, URetinex \cite{wu2022uretinexnet}, and SCI \cite{ma2022sci}. The comparison  demonstrates that the SCI method exhibits significant advantages overall when processing the LLOT dataset: enhanced images display clearer object edge contours; image noise is relatively less pronounced; and the enhancement effect more closely resembles natural human visual perception. Fig. \ref{fig:enhanceAl} presents visual comparison examples of these algorithms. Furthermore, SCI-processed images do not consume excessive memory, which is particularly important for practical applications. Considering these factors comprehensively, the SCI algorithm demonstrates clear advantages in both visual quality and practicality, making it the optimal choice for LLOT dataset preprocessing. Our annotation process follows the guidelines proposed in \cite{muller2018trackingnet} and involves two main aspects: annotating visible objects and marking occlusions. When the target is visible, a bounding box aligned with the coordinate axes is drawn or edited for each frame in the sequence, starting from the given initial target, ensuring the box tightly encloses all visible parts of the target. When the target is not visible, it is marked as out of view (OV), partially occluded (POC), or fully occluded (FOC). To ensure the high quality and reliability of the LLOT, we employed a rigorous three-step annotation process. First, each video was initially annotated by a professional (a student researching visual tracking). Then, the validation team conducted a thorough visual inspection of these initial annotations, focusing on their accuracy and consistency. Finally, annotations disputed by the validation team were returned to the original annotator for bounding box correction and refinement.
This annotation approach not only enhances the overall quality of the LLOT but also ensures consistency and accuracy in the annotations. Fig .\ref{fig:lable} shows examples of box annotations in the LLOT.

\subsection{Attributes}
\begin{figure}[!t]
\centering
\includegraphics[width=3.6in]{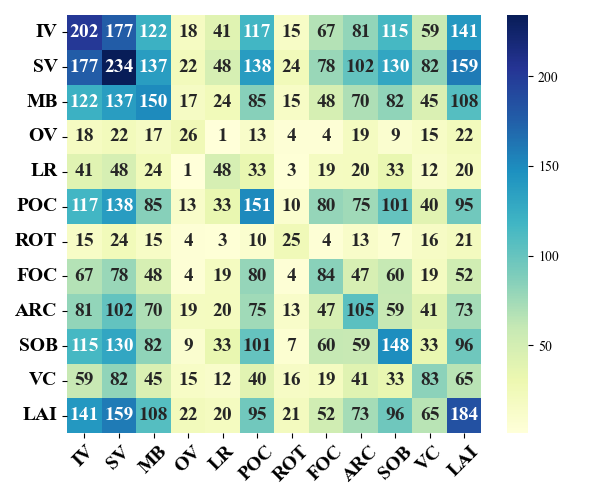}
\caption{ Attribute distribution of LLOT benchmark. The values along the matrix diagonal reflect the distribution across the entire benchmark, while each row or column depicts the joint distribution of a specific attribute subset.}
\label{fig:att_matrix}
\end{figure}

To further analyze tracking performance, the LLOT provides 11 common attributes for each sequence: illumination variation (IV), scale variation (SV), motion blur (MB), out of view (OV), partial occlusion (POC), rotation (ROT), full occlusion (FOC), viewpoint change (VC), similar objects (SOB), aspect ratio change (ARC), and low resolution (LR). Additionally, there is an attribute related to low-light conditions—low ambient intensity (LAI) \cite{ye2022nat2021}. Table \ref{tab:attr_meanings} details the definition of each attribute. Notably, the ARC, SV, LR, and LAI attributes are computed from the annotation results of the targets, while the remaining attributes are manually labeled. 
Overall, Fig. \ref{fig:att_matrix} shows a heatmap illustrating the relationships between each attribute. Attribute pairs that frequently appear together are represented by darker colors, while less common combinations are depicted in lighter shades. The values along the matrix diagonal specifically reflect the distribution across the entire benchmark, while each row or column depicts the joint distribution of a specific attribute subset. We observe that IV, SV, and LAI are the most prevalent in the LLOT dataset. Following these, MB, POC, and SOB also exhibit relatively high occurrence frequencies. Among these, IV, SV, MB, POC and SOB are common challenges in traditional tracking tasks, while LAI is a new attribute specifically designed in \cite{ye2022nat2021} to study the impact of illumination on trackers.

\section{A New Baseline: H-DCPT}

In our evaluation, we find that HIPTrack \cite{cai2024hiptrack} significantly improves tracking performance by utilizing precise and updated historical prompt information, outperforming other state-of-the-art trackers on the LLOT dataset. However, HIPTrack's performance still has significant potential for further improvement, as it does not specifically address the unique challenges of low-light conditions. 
On the other hand, DCPT \cite{zhu2023dcpt}, a state-of-the-art tracker designed for nighttime scenarios, enhances its performance in low-light conditions using darkness clue prompts (DCP). However, it fails to adequately leverage essential target information from previous frames, leading to suboptimal overall performance on the LLOT dataset.
Based on these observations, we propose an innovative approach to integrate the strengths of both trackers. Specifically, we apply the DCP module of DCPT within the HIPTrack framework. This aims to combine HIPTrack's advantage of leveraging historical prompt information with DCPT's expertise in handling low-light environments. This hybrid approach allows HIPTrack to maintain its efficient use of historical data while significantly enhancing its performance under low-light conditions. Through this integration, we aim to develop a tracking algorithm that maintains high performance across various lighting conditions, particularly excelling in low-light environments.

\begin{figure*}[!t]
\centering
\includegraphics[width=1\textwidth]{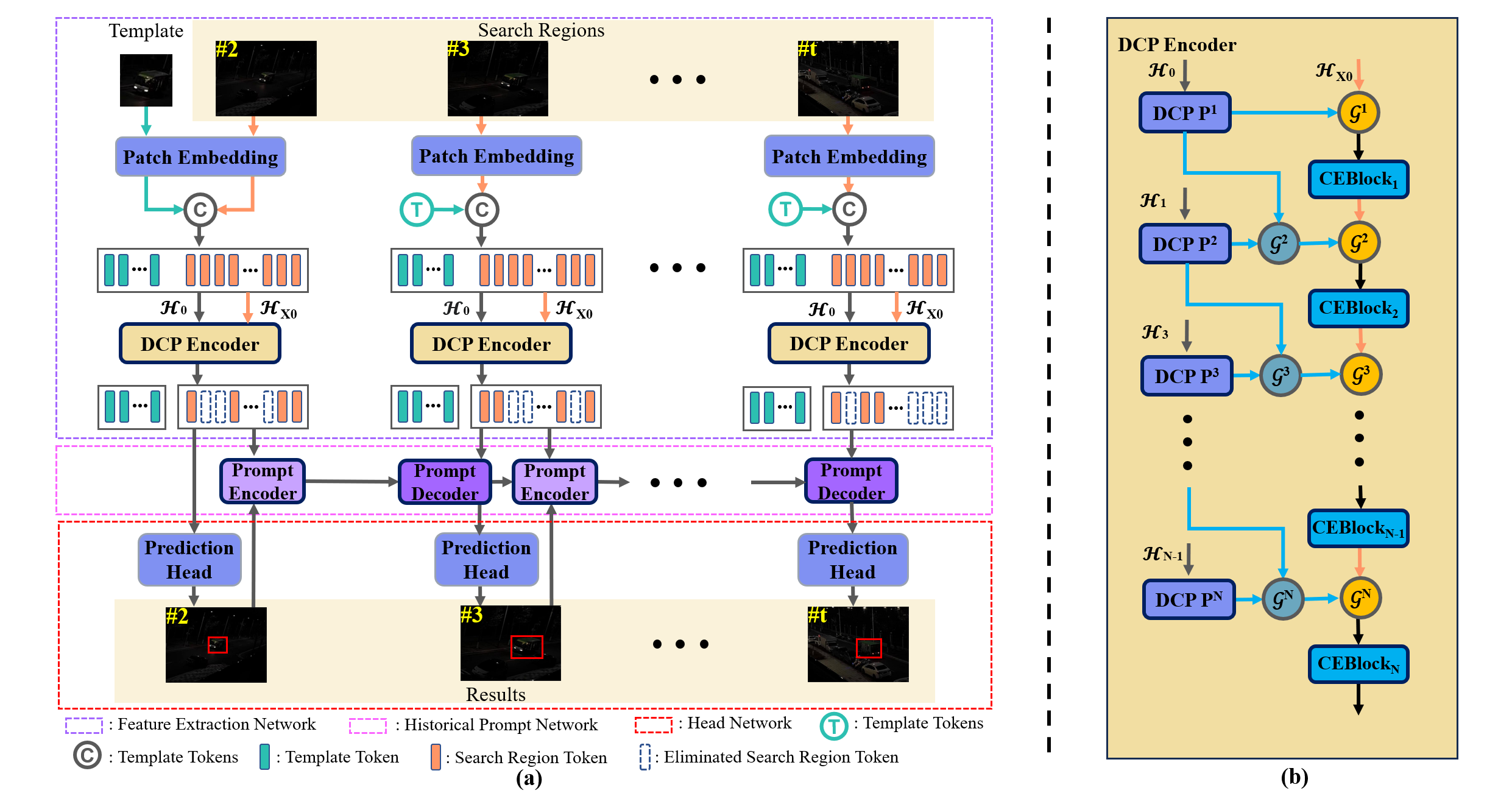}
\caption{(a) An illustration of the proposed H-DCPT method. The network structure is inherited from that of HIPTrack. The difference lies in the encoder part of the feature extraction network. (b) The structure of the DCP Encoder}
\label{H-DCPT}
\vspace{-0.15in}
\end{figure*}

\subsection{Overall Architecture}
H-DCPT builds upon the fundamental architectures of HIPTrack, comprising three core components as illustrated in Fig. \ref{H-DCPT}(a). The feature extraction network in H-DCPT serves two primary functions: 1) it extracts target-template matching information from the search area while effectively filtering out background interference; 2) H-DCPT introduces a specialized DCP module within the feature extraction network, specifically designed for low-light environments. By integrating darkness clue prompting techniques, H-DCPT significantly enhances its feature representation capabilities in low-light conditions while retaining the ability to process historical prompt information. The historical prompt network consists of a historical prompt encoder and decoder. The encoder transforms the current target state into historical features for storage, while the decoder utilizes this information to generate corresponding historical prompts for the search area, subsequently fusing them with compressed features. H-DCPT retains the head network structure of HIPTrack. The distinction between H-DCPT and HIPTrack lies in their feature extraction networks. HIPTrack employs a Vision Transformer (ViT) architecture incorporating an early candidate elimination module, similar to the approach used in OSTrack \cite{ye2022ostrack}, as its feature encoder. In contrast, our H-DCPT introduces a novel aggregation encoder, as shown in Fig. \ref{H-DCPT}(b),  which adaptively aggregates darkness clue prompts and base features through a Gated Feature Aggregation (GFA) mechanism. Through this design, H-DCPT effectively combines the advantages of HIPTrack’s utilization of historical prompt information with DCPT's \cite{zhu2023dcpt} adaptability to low-light conditions, achieving efficient tracking across various lighting environments, especially excelling in low-light enviroment. In subsequent section, we provides a detailed overview of the proposed DCP Encoder. Other architectural details remain consistent with HIPTrack, and further information can refer to \cite{cai2024hiptrack}.

\subsection{DCP encoder}
We developed our DCP encoder based on the Transformer architecture from HIPTrack. Specifically, it adopts the Vision Transformer (ViT) framework and additionally incorporates an early candidate elimination module. This structure is represented by stacked CEBlocks, as shown in Fig. \ref{H-DCPT}(b). The primary distinction is the addition of both the DCP module and the Gated Feature Aggregation (GFA) strategy. Specifically, we position an independent DCP module before each CEBlock. These modules compute and generate darkness clue prompt information. Subsequently, similar to DCPT, our encoder utilizes the GFA mechanism to dynamically adjust the weights between features generated by CEBlocks and darkness clue prompts. This design enables the model to more flexibly integrate information from different sources, thereby significantly enhancing its performance under low-light conditions.

 The process of computing the darkness clue prompts injected into the i-th feature gate can be expressed as:
\begin{equation}
\mathcal{P}^i = (\mathcal{G}^i)\Phi(\mathcal{H}^{i-1}) + (1-\mathcal{G}^i )\mathcal{P}^{i-1},
\end{equation}
where $\Phi$ denotes the DCP extraction operation, and $\mathcal{H}^{i-1}$ is the template and search features concatenated at the i-th layer. The gating weight $\mathcal{G}^i$ controls the combination of the current and previous prompts. The effect of the proposed DCP encoder can be formalized as:
\begin{equation}
F_{ha}= \mathcal{H}_{xn} + \rho^n \mathcal{P}^n,
\end{equation}
where $\mathcal{H}_{xn}=CBlock_n(\mathcal{H}_{n-1})$, and the gated aggregation weight $\rho^n$ represents the varying attention weights of the darkness clue prompts at the n-th layer.

\begin{table*}
\centering
\setlength{\tabcolsep}{10pt} 
\renewcommand{\arraystretch}{1.2} 

\caption{The overall performance of baseline trackers on the LLOT dataset. The trackers are ranked in descending order based on their S$_{AUC}$ scores, with the top three performers highlighted in \textcolor{red}{\textbf{red}}, \textcolor{blue}{\textbf{blue}}, and \textcolor{green}{\textbf{green}}, respectively. The Properties column denotes the attributes of different trackers, categorized into three aspects: correlation filters (Yes/No), deep learning (Yes/No), and feature representation, which includes Transformer (Trans), Convolutional Neural Networks (CNN), fast Histogram of Gradients (fHOG), Color Names (CN), Color Histograms (CH), and Natural Language (NL).}
\begin{tabular}{clcccclcccc} 
\toprule
\multicolumn{1}{l}{\multirow{2}{*}{\textbf{Rank}}} & \multirow{2}{*}{\textbf{Tracker }} & \multicolumn{3}{c}{\textbf{Performance}}                                                                                          &  & \multicolumn{4}{c}{\textbf{Properties}}       & \multirow{2}{*}{\textbf{ Venue}}  \\ 
\cline{3-5}\cline{7-10}
\multicolumn{1}{l}{}                               &                                    & \textbf{S$_{AUC}$}                          & \textbf{P}                            & \textbf{P$_{Norm}$}                        &  & \textbf{CF} & \textbf{DL} & \textbf{Repr.} &  &                                   \\ 
\hline
\textbf{1}   &H-DCPT                             & \textcolor{red}{\textbf{0.576}}            & \textcolor{red}{\textbf{0.684}}            & \textcolor{red}{\textbf{0.739}}            &  &                               & \CheckmarkBold & CNN+Trans      &  & Ours                              \\
\textbf{2}   &HIPTrack \cite{cai2024hiptrack}                           & \textcolor{blue}{\textbf{0.560}}           & \textcolor{blue}{\textbf{0.662}}           & \textbf{\textcolor{blue}{0.716}}           &  &                               & \CheckmarkBold & Trans          &  & CVPR2024                          \\
\textbf{3}   &ProContEXT \cite{lan2023procontext}                         & \textcolor{green}{\textbf{0.557}}  & \textcolor{green}{\textbf{0.660}} & \textcolor{green}{\textbf{0.708}} &  &                               & \CheckmarkBold & Trans          &  & CVPR2023                          \\
\textbf{4}   &ARTrack \cite{wei2023artrack}                           & 0.553                                      & 0.656                                      & 0.707                                      &  &                               & \CheckmarkBold & Trans          &  & CVPR2023                          \\
\textbf{5}   &DropTrack \cite{wu2023droptrack}                         & 0.549                                      & 0.648                                      & 0.701                                      &  &                               & \CheckmarkBold & Trans          &  & CVPR2023                          \\
\textbf{6}   &AQATrack \cite{xie2024aqatrack}                          & 0.549                                      & 0.646                                      & 0.699                                      &  &                               & \CheckmarkBold & Trans          &  & CVPR2024                          \\
\textbf{7}   &ROMTrack \cite{cai2023romtrack}                          & 0.540                                      & 0.650                                      & 0.698                                      &  &                               & \CheckmarkBold & Trans          &  & ICCV2023                          \\
\textbf{8}   &DCPT \cite{zhu2023dcpt}                              & 0.527                                      & 0.614                                      & 0.665                                      &  &                               & \CheckmarkBold & CNN+Trans      &  & ICRA2024                          \\
\textbf{9}   &OSTrack \cite{ye2022ostrack}                           & 0.521                                      & 0.613                                      & 0.663                                      &  &                               & \CheckmarkBold & Trans          &  & ECCV2022                          \\
\textbf{10}  &GRM \cite{gao2023grm}                               & 0.516                                      & 0.609                                      & 0.660                                      &  &                               & \CheckmarkBold & Trans          &  & CVPR2023                          \\
\textbf{11}  &STARK \cite{yan2021stark}                             & 0.516                                      & 0.599                                      & 0.644                                      &  &                               & \CheckmarkBold & CNN+Trans      &  & ICCV2021                          \\
\textbf{12}  &SeqTrack \cite{chen2023seqtrack}                          & 0.514                                      & 0.615                                      & 0.673                                      &  &                               & \CheckmarkBold & Trans          &  & CVPR2023                          \\
\textbf{13}  &ZoomTrack \cite{kou2024zoomtrack}                         & 0.498                                      & 0.594                                      & 0.641                                      &  &                               & \CheckmarkBold & Trans          &  & NIPS2023                          \\
\textbf{14}  &SimTrack \cite{chen2022simtrack}                          & 0.493                                      & 0.574                                      & 0.620                                      &  &                               & \CheckmarkBold & Trans          &  & ECCV2022                          \\
\textbf{15}  &CSWinTT \cite{song2022cswintt}                           & 0.466                                      & 0.533                                      & 0.585                                      &  &                               & \CheckmarkBold & CNN+Trans      &  & CVPR2022                          \\
\textbf{16}  &STRCF \cite{li2018strcf}                             & 0.459                                      & 0.555                                      & 0.602                                      &  & \CheckmarkBold &                               & fHOG+CN+GS     &  & CVPR2018                          \\
\textbf{17}  &SRDCFdecon \cite{2016srdcfdecon}                        & 0.456                                      & 0.563                                      & 0.598                                      &  & \CheckmarkBold &                               & fHOG           &  & CVPR2016                          \\
\textbf{18}  &SRDCF \cite{danelljan2015srdcf}                             & 0.448                                      & 0.555                                      & 0.586                                      &  & \CheckmarkBold &                               & fHOG           &  & ICCV2015                          \\
\textbf{19}  &BACF \cite{kiani2017bacf}                              & 0.444                                      & 0.552                                      & 0.593                                      &  & \CheckmarkBold &                               & fHOG           &  & ICCV2017                          \\
\textbf{20}  &ECO-HC \cite{danelljan2017eco_hc}                            & 0.441                                      & 0.542                                      & 0.568                                      &  & \CheckmarkBold &                               & fHOG+CN        &  & CVPR2017                          \\
\textbf{21}  &ARCF \cite{huang2019arcf}                              & 0.43.6                                      & 0.540                                      & 0.585                                      &  & \CheckmarkBold &                               & fHOG           &  & CVPR2019                          \\
\textbf{22}  &AutoTrack \cite{li2020autotrack}                         & 0.430                                      & 0.538                                      & 0.571                                      &  & \CheckmarkBold &                               & fHOG+CN+GS     &  & CVPR2020                          \\
\textbf{23}  &Aba-ViTrack \cite{li2023aba-vitrack}                       & 0.430                                      & 0.504                                      & 0.545                                      &  &                               & \CheckmarkBold & Trans          &  & ICCV2023                          \\
\textbf{24}  &Staple\_CA \cite{mueller2017staple_ca}                        & 0.417                                      & 0.511                                      & 0.543                                      &  & \CheckmarkBold &                               & fHOG+CN        &  & CVPR2017                          \\
\textbf{25}  &UDAT \cite{ye2022udat}                              & 0.409                                      & 0.513                                      & 0.538                                      &  &                               & \CheckmarkBold & CNN+Trans      &  & CVPR2022                          \\
\textbf{26}  &Staple \cite{bertinetto2016staple}                            & 0.408                                      & 0.517                                      & 0.541                                      &  & \CheckmarkBold &                               & fHOG+CH        &  & CVPR2016                          \\
\textbf{27}  &SiamGAT \cite{guo2021siamgat}                           & 0.406                                      & 0.491                                      & 0.531                                      &  &                               & \CheckmarkBold & CNN            &  & CVPR2021                          \\
\textbf{28}  &AVTrack \cite{li2024avtrack}                           & 0.403                                      & 0.468                                      & 0.508                                      &  &                               & \CheckmarkBold & Trans          &  & ICML2024                          \\
\textbf{29}  &SAM-DA \cite{fu2023samda}                            & 0.396                                      & 0.502                                      & 0.528                                      &  &                               & \CheckmarkBold & CNN+Trans      &  & ArXiv2023                         \\
\textbf{30}  &DSST \cite{danelljan2014dsst}                              & 0.379                                      & 0.471                                      & 0.499                                      &  & \CheckmarkBold &                               & fHOG           &  & BMVC2014                          \\
\textbf{31}  &KCC \cite{wang2018kcc}                               & 0.369                                      & 0.495                                      & 0.521                                      &  & \CheckmarkBold &                               & fHOG+CN        &  & AAAI2018                          \\
\textbf{32}  &SAMF \cite{li2015samf}                              & 0.368                                      & 0.492                                      & 0.504                                      &  & \CheckmarkBold &                               & fHOG+CN        &  & ECCV2014                          \\
\textbf{33}  &ETTrack \cite{blatter2023ettrack}                           & 0.358                                      & 0.425                                      & 0.446                                      &  &                               & \CheckmarkBold & Trans      &  & WACV2023                          \\
\textbf{34}  &fDSST \cite{danelljan2016fdsst}                             & 0.341                                      & 0.429                                      & 0.438                                      &  & \CheckmarkBold &                               & fHOG+GS        &  & TPAMI2017                         \\
\textbf{35}  &JointNLT \cite{zhou2023jointnlt}                          & 0.328                                      & 0.356                                      & 0.401                                      &  &                               & \CheckmarkBold & NL+Trans       &  & CVPR2023                          \\
\textbf{36}  &KCF \cite{henriques2014kcf}                               & 0.325                                      & 0.441                                      & 0.462                                      &  & \CheckmarkBold &                               & fHOG           &  & TPAMI2015                         \\
\textbf{37}  &SAMF\_CA \cite{mueller2017samf_ca}                          & 0.315                                      & 0.429                                      & 0.426                                      &  & \CheckmarkBold &                               & fHOG+CN        &  & CVPR2017                          \\
\textbf{38}  &HiFT \cite{cao2021hift}                              & 0.268                                      & 0.345                                      & 0.358                                      &  &                               & \CheckmarkBold & CNN            &  & ICCV2021                          \\
\textbf{39}  &TCTrack \cite{cao2022tctrack}                           & 0.252                                      & 0.353                                      & 0.363                                      &  &                               & \CheckmarkBold & CNN+Trans      &  & CVPR2022                          \\
\textbf{40}  &SiamAPN++ \cite{cao2021siamapn++}                         & 0.216                                      & 0.300                                      & 0.285                                      &  &                               & \CheckmarkBold & CNN            &  & IROS2021                          \\
\bottomrule
\end{tabular}
\setlength{\tabcolsep}{6pt}
\renewcommand{\arraystretch}{1} 
\label{tab:overallperformance}
\end{table*}

\section{Evaluation}

In this study, we employ three widely used quantitative metrics to evaluate tracker performance: Success (S$_{AUC}$), Precision (P), and Normalized Precision (P$_{Norm}$). Using the standard One-Pass Evaluation (OPE) protocol \cite{wu2015otb}, we measure these metrics across 269 sequences. S evaluates performance based on bounding box overlap. The overlap score, defined as $S=\frac{\left|B_{tr} \cap B_{gt}\right|}{\left|B_{tr} \cup B_{gt}\right|}$, where $B_{tr}$ is the tracking result and $B_{gt}$ is the ground truth, measures the ratio of successful frames across overlap thresholds from 0 to 1. Using a single success rate value at a specific threshold (e.g., $\tau$ = 0.5) may not be fair or representative for evaluating trackers. Therefore, we rank overall performance using the area under the curve (AUC) of this plot, denoted as S$_{AUC}$, which provides a more comprehensive assessment across all thresholds. $P=||C_{tr} - C_{gt}||$ assesses tracking accuracy by measuring the Euclidean distance between the tracking results (C$_{tr}$) and ground truth target annotation (C$_{gt}$). To mitigate potential biases when trackers lose targets, we use a 20-pixel distance threshold, following established practices \cite{babenko2010robust}. And P$_{Norm}$ is a scale-invariant metric that normalizes precision relative to the ground truth size, formulated as $P_{Norm}=||diag(B_{tr},B_{gt})(C_{tr} - C_{gt})||$. This metric is particularly robust across varying object sizes and scales, and trackers are ranked using the AUC calculated between 0 and 0.5. These metrics together provide a comprehensive assessment, enabling accurate comparison and evaluation of different tracking algorithms.

\subsection{Trackers for Comparison}
We evaluate our H-DCPT along with 39 state-of-the-art trackers to understand their performance on LLOT. These trackers are categorized into two groups: 24 deep learning based (DL-based) trackers and 15 Correlation filters based (DCF-based) trackers. Specifically, the DL-based trackers include HIPTrack \cite{cai2024hiptrack}, ProContEXT \cite{lan2023procontext}, ARTrack \cite{wei2023artrack}, DropTrack \cite{wu2023droptrack}, AQATrack \cite{xie2024aqatrack}, ROMTrack \cite{cai2023romtrack}, DCPT \cite{zhu2023dcpt}, OStrack \cite{ye2022ostrack}, GRM \cite{gao2023grm}, Stark \cite{yan2021stark}, SeqTrack \cite{chen2023seqtrack}, ZoomTrack \cite{kou2024zoomtrack}, SimTrack \cite{chen2022simtrack}, CSWinTT \cite{song2022cswintt}, Aba-ViTrack \cite{li2023aba-vitrack}, UDAT \cite{ye2022udat}, SiamGAT \cite{guo2021siamgat}, AVTrack \cite{li2024avtrack}, SAM-DA \cite{fu2023samda}, ETTrack \cite{blatter2023ettrack}, JointNLT \cite{zhou2023jointnlt}, HiFT \cite{cao2021hift}, TCTrack \cite{cao2022tctrack}, and SiamAPN++ \cite{cao2021siamapn++}. The DCF-based trackers comprise STRCF \cite{li2018strcf}, SRDCFdecon \cite{2016srdcfdecon}, SRDCF \cite{danelljan2015srdcf}, BACF \cite{kiani2017bacf}, ECO-HC \cite{danelljan2017eco_hc}, ARCF \cite{huang2019arcf}, AutoTrack \cite{li2020autotrack}, Staple\_CA \cite{mueller2017staple_ca}, Staple \cite{bertinetto2016staple}, DSST \cite{danelljan2014dsst}, KCC \cite{wang2018kcc}, SAMF \cite{li2015samf}, fDSST \cite{danelljan2016fdsst}, KCF \cite{henriques2014kcf}, and SAMF\_CA \cite{mueller2017samf_ca}. Notably, the code library for the DCF-based trackers is sourced from \cite{fu2021correlation}. To ensure consistency and fairness in our evaluation, we utilize the original implementations provided by the authors for all trackers, including their pre-trained models and recommended parameter settings. This comprehensive comparison allows us to assess the effectiveness of our proposed H-DCPT against a wide range of modern tracking algorithms in the challenging LLOT scenario.

\subsection{Evaluation Results}
\noindent\textbf{Overall Performance:}
A comprehensive evaluation of 39 state-of-the-art trackers and our H-DCPT was conducted on the LLOT dataset. Table \ref{tab:overallperformance} illustrates the evaluation results of all the trackers, ranked by their S$_{AUC}$ scores. Overall, we observe that the top 15 ranked trackers all employ DL-based methods. Trackers in the mid-range rankings demonstrate a diversity of approaches, including both DCF-based methods and DL-based techniques. Notably, the bottom three trackers are also DL-based. This suggests that DL-based methods generally hold significant advantages in low-light object tracking, due to their powerful feature extraction and representation capabilities, which enable better adaptation to complex low-light scenarios.  
For example, among the 40 trackers evaluated, which include 39 existing advanced trackers and the newly proposed H-DCPT, experimental results indicate that H-DCPT, HIPTrack \cite{cai2024hiptrack}, and ProcontEXT \cite{lan2023procontext} achieve the top three performances. While HIPTrack and ProcontEXT primarily rely on Transformer architectures for feature representation, H-DCPT distinguishes itself by leveraging a hybrid approach that combines Transformers with CNN. H-DCPT achieves the best performance with scores of 0.576, 0.684, and 0.739, outperforms the second-placed HIPTrack by 1.6\%, 2.2\%, and 2.3\% for S$_{AUC}$, P, and P$_{Norm}$ respectively. HIPTrack ranking second, with scores of 0.560, 0.662, and 0.716 in the corresponding metrics. ProContEXT follows closely in third place, achieving 0.557, 0.660, and 0.708 for S$_{AUC}$, P, and P$_{Norm}$. 
However, some DL-based algorithms that perform exceptionally well on daytime datasets do not outperform certain DCF-based trackers on the LLOT benchmark. 
Among the traditional trackers utilizing only handcrafted features, STRCF \cite{li2018strcf}, SRDCFdecon \cite{2016srdcfdecon}, SRDCF \cite{danelljan2015srdcf}, BACF \cite{kiani2017bacf}, and ECO-HC \cite{danelljan2017eco_hc} achieve the top five rankings in performance evaluation. Notably, despite not employing deep learning features, these methods demonstrate overall performance comparable to some deep learning trackers, such as CSWinTT \cite{song2022cswintt}, across three evaluation metrics. Moreover, they even outperform trackers specifically designed for nighttime tracking, including UDAT \cite{ye2022udat} and SAM-DA \cite{fu2023samda}. Among the 39 state-of-the-art trackers evaluated, the DL-based algorithms HiFT \cite{cao2021hift}, TCTrack \cite{cao2022tctrack}, and SiamAPN++ \cite{cao2021siamapn++} demonstrate relatively weak performance on the low-light dataset LLOT. Specifically, HiFT achievs scores of 0.268, 0.345, and 0.358 in Success (S$_{AUC}$), Precision (P), and Normalized Precision (P$_{Norm}$), respectively. TCTrack obtains scores of 0.252, 0.353, and 0.363 on the same metrics. SiamAPN++ shows the poorest performance, with scores of only 0.216, 0.300, and 0.285. One possible reason for such situation may be that all three trackers are designed for drone scenarios. To maintain real-time tracking capabilities, these algorithms might sacrifice robustness under complex lighting conditions, which explains their suboptimal performance in low-light environments.

In conclusion, the experimental results demonstrate that H-DCPT exhibits superior performance in terms of Success (S$_{AUC}$), Precision (P), and Normalized Precision (P$_{Norm}$). This confirms that H-DCPT, by integrating dark cue prompts and historical prompt information, shows significant effectiveness in addressing the challenges of low-light object tracking compared to existing state-of-the-art trackers. These findings also reveal a critical issue: previous tracking methods have primarily focused on robust tracking in well-lit scenarios while neglecting performance in low-light environments. Under these more complex and demanding conditions, these trackers often face risks of performance degradation or even failure. As the demand for low-light tracking increases, research and innovation in this field will become increasingly crucial. Developing robust tracking algorithms capable of maintaining high performance across various lighting conditions will be a key challenge and opportunity in the fields of computer vision and object tracking.

\begin{figure*}[t]
    \centering
    \begin{minipage}[t]{0.32\textwidth}
        \centering
        \includegraphics[width=1\textwidth]{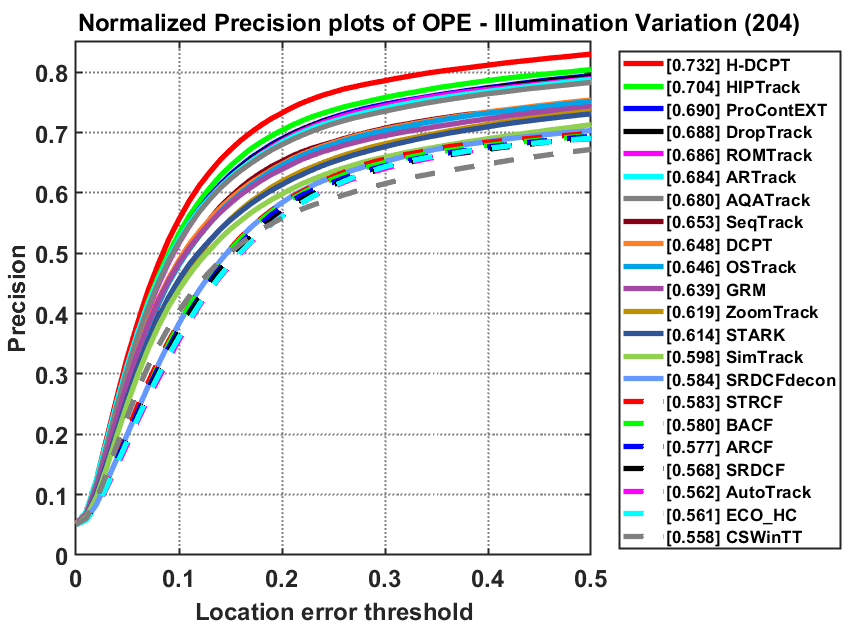}\hspace{0in}
    \end{minipage}
    \begin{minipage}[t]{0.32\textwidth}
        \centering
        \includegraphics[width=1\textwidth]{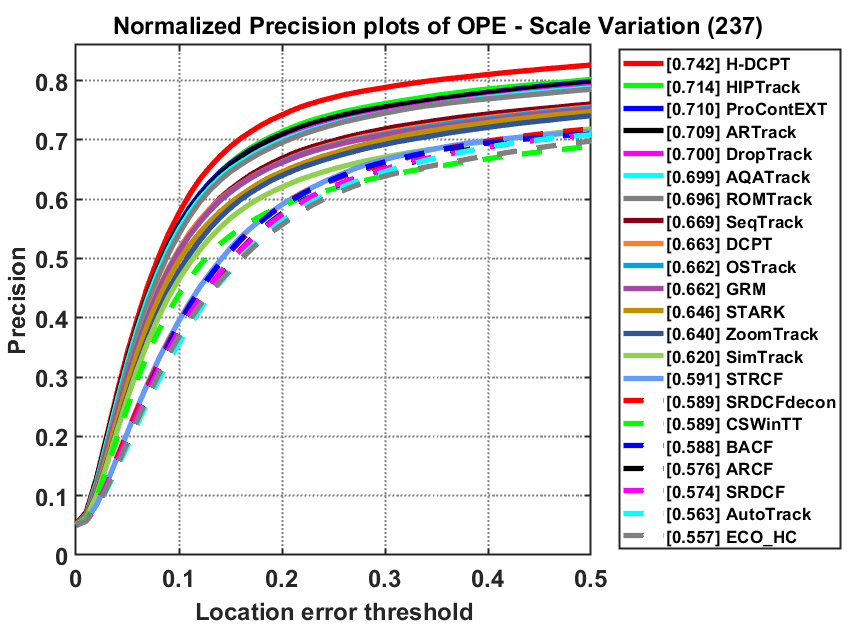}\hspace{0in}	
    \end{minipage}
    \begin{minipage}[t]{0.32\textwidth}
        \centering
        \includegraphics[width=1\textwidth]{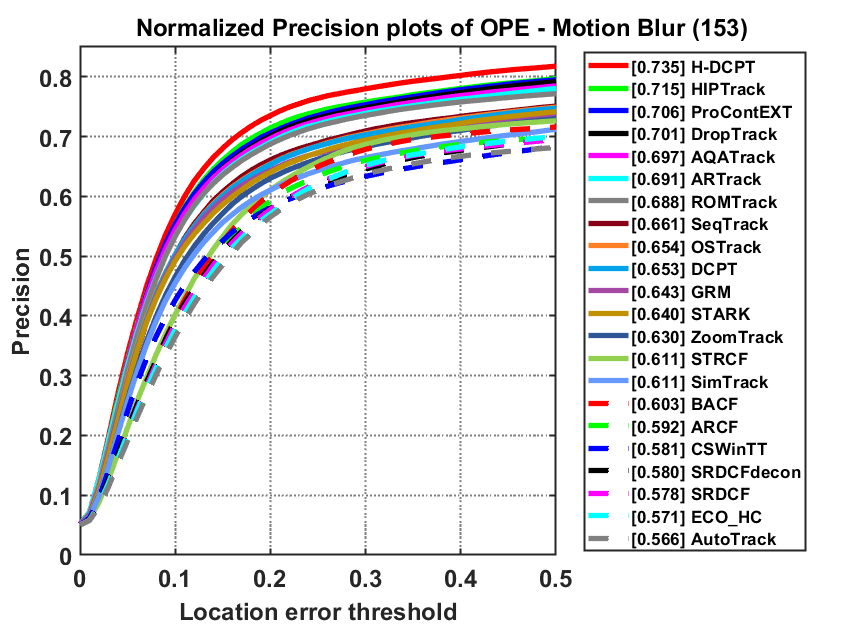}\hspace{0in}	
    \end{minipage}
    \vspace{-0.1in}

    \begin{minipage}[t]{0.32\textwidth}
        \centering
        \includegraphics[width=1\textwidth]{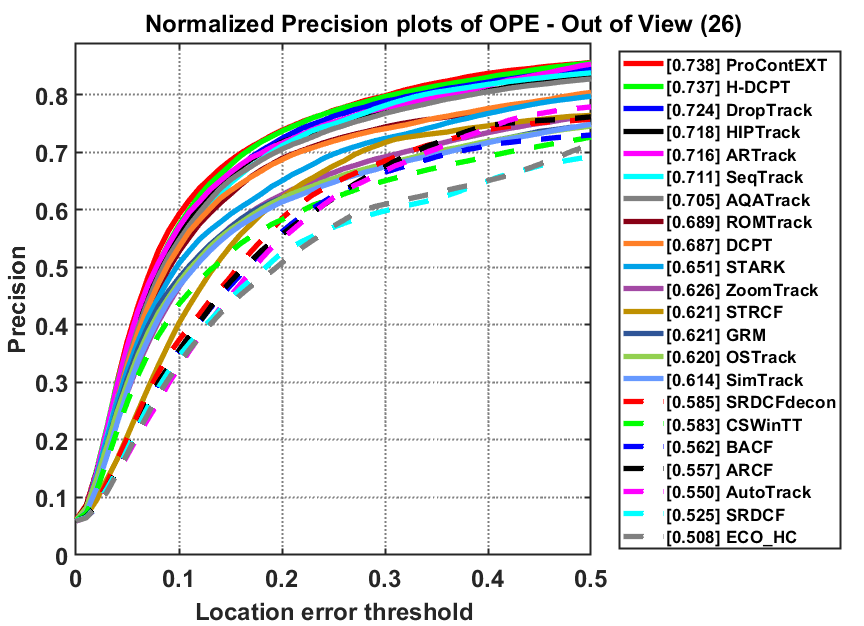}\hspace{0in}
    \end{minipage}
    \begin{minipage}[t]{0.32\textwidth}
        \centering
        \includegraphics[width=1\textwidth]{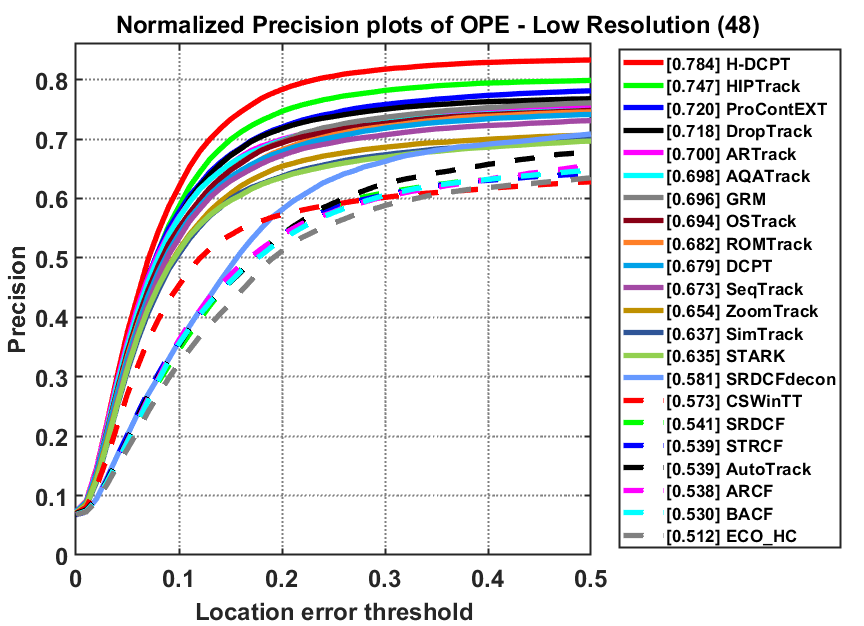}\hspace{0in}	
    \end{minipage}
    \begin{minipage}[t]{0.32\textwidth}
        \centering
        \includegraphics[width=1\textwidth]{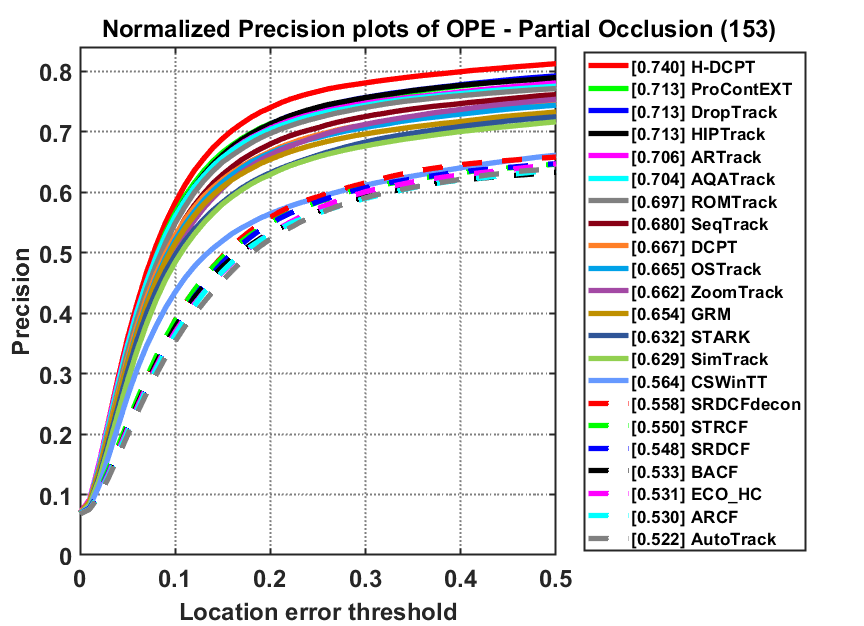}\hspace{0in}	
    \end{minipage}
    \vspace{-0.1in}

        \begin{minipage}[t]{0.32\textwidth}
        \centering
        \includegraphics[width=1\textwidth]{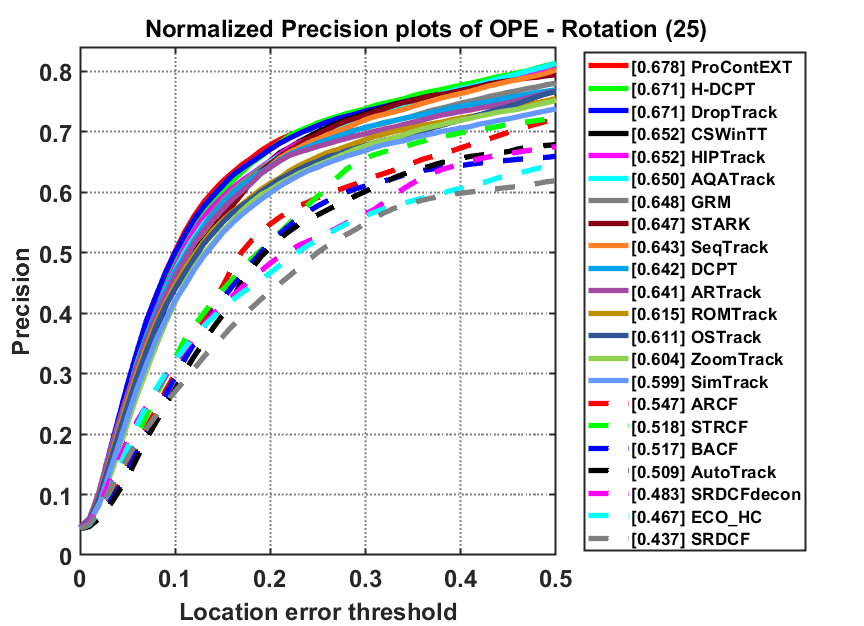}\hspace{0in}
    \end{minipage}
    \begin{minipage}[t]{0.32\textwidth}
        \centering
        \includegraphics[width=1\textwidth]{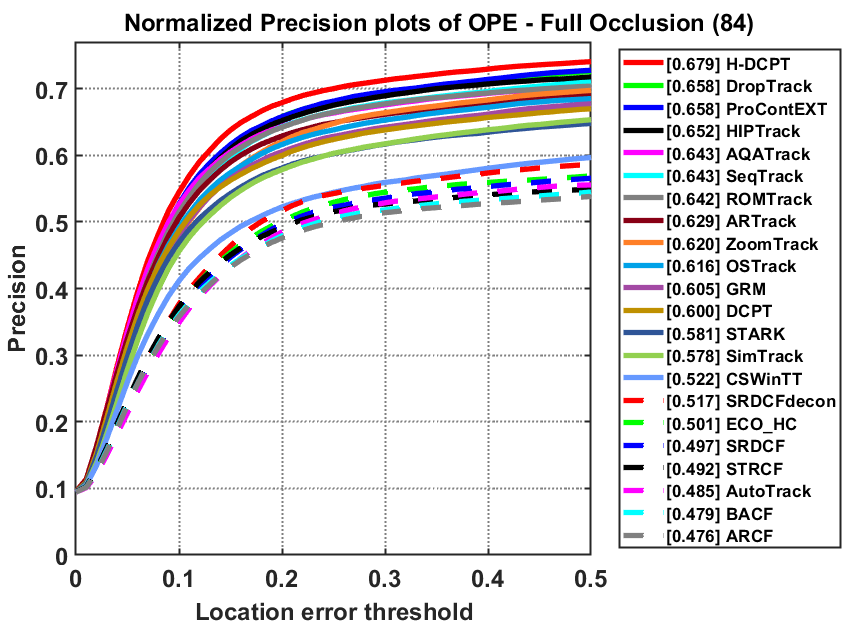}\hspace{0in}	
    \end{minipage}
    \begin{minipage}[t]{0.32\textwidth}
        \centering
        \includegraphics[width=1\textwidth]{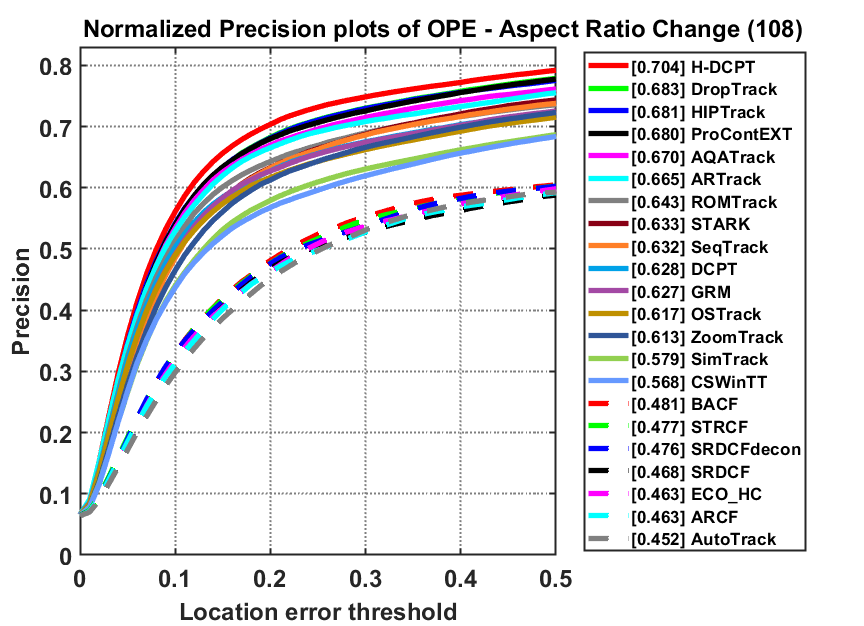}\hspace{0in}	
    \end{minipage}
    \vspace{-0.1in}

        \begin{minipage}[t]{0.32\textwidth}
        \centering
        \includegraphics[width=1\textwidth]{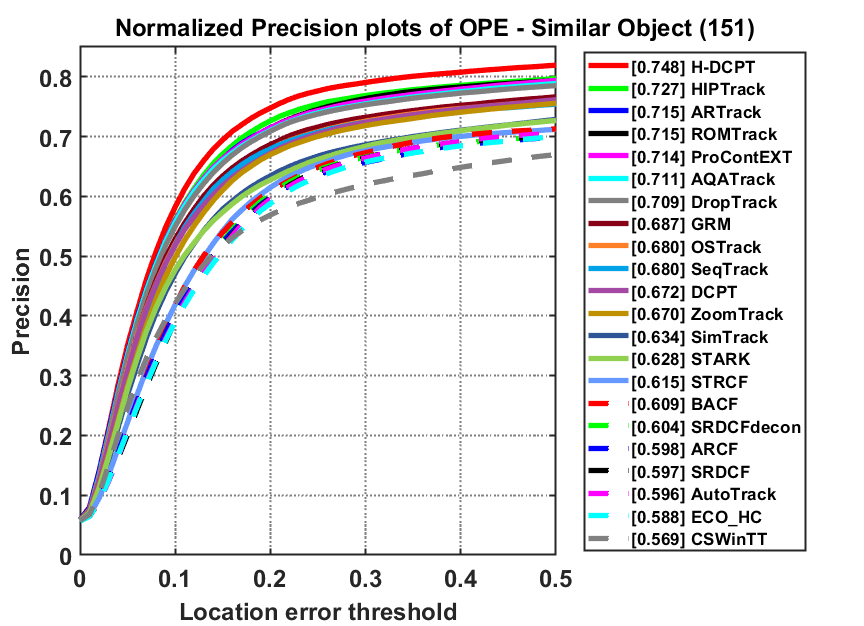}\hspace{0in}
    \end{minipage}
    \begin{minipage}[t]{0.32\textwidth}
        \centering
        \includegraphics[width=1\textwidth]{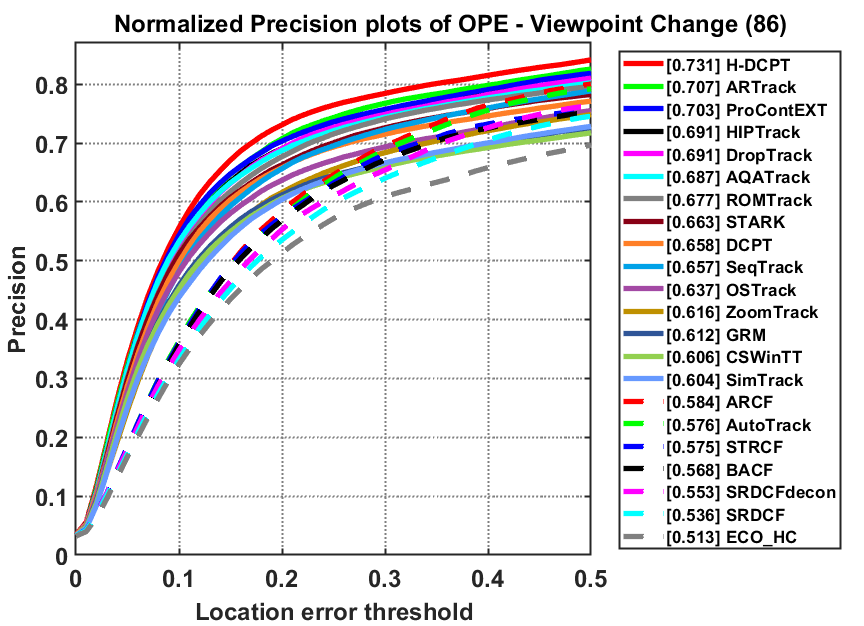}\hspace{0in}	
    \end{minipage}
    \begin{minipage}[t]{0.32\textwidth}
        \centering
        \includegraphics[width=1\textwidth]{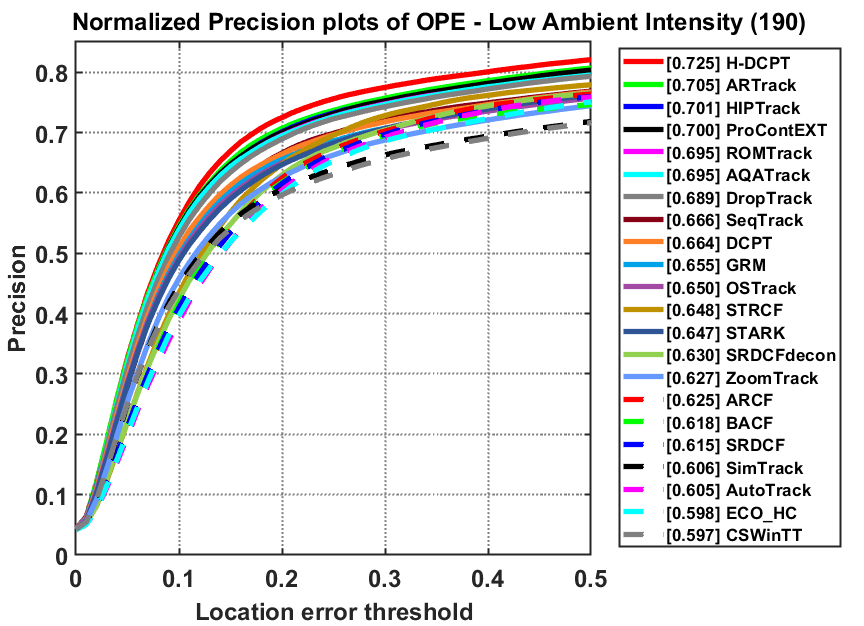}\hspace{0in}	
    \end{minipage}

    \caption{The Normalized Precision performance of trackers on each attribute under the one-pass evaluation (OPE) \cite{wu2015otb} protocol. For clarity, only the top 22 performing trackers are displayed.}
    \label{fig:LLOT_Att_OPE}
    \vspace{-0.15in}
\end{figure*}

\noindent\textbf{Attribute-based performance:} We conduct a comprehensive performance evaluation across 12 attributes to gain deeper insights into the performance of different trackers on the LLOT dataset. Fig. \ref{fig:LLOT_Att_OPE} illustrates the Normalized Precision (P$_{Norm}$) performance of each tracker under the One-Pass Evaluation (OPE) protocol \cite{wu2015otb}. To improve readability, only the top 22 trackers are displayed in the figure, including 21 state-of-the-art trackers (HIPTrack \cite{cai2024hiptrack}, ProContEXT \cite{lan2023procontext}, DropTrack \cite{wu2023droptrack}, ROMTrack \cite{cai2023romtrack}, ARTrack \cite{wei2023artrack}, AQATrack \cite{xie2024aqatrack}, SeqTrack \cite{chen2023seqtrack}, DCPT \cite{zhu2023dcpt}, OStrack \cite{ye2022ostrack}, GRM \cite{gao2023grm}, ZoomTrack \cite{kou2024zoomtrack}, Stark \cite{yan2021stark}, SimTrack \cite{chen2022simtrack}, CSWinTT \cite{song2022cswintt}, STRCF \cite{li2018strcf}, SRDCFdecon \cite{2016srdcfdecon}, SRDCF \cite{danelljan2015srdcf}, BACF \cite{kiani2017bacf}, ECO-HC \cite{danelljan2017eco_hc}, ARCF \cite{huang2019arcf}, and AutoTrack \cite{li2020autotrack}) and our proposed H-DCPT. The attribute plots illustrate that H-DCPT performs the best in 10 out of the 12 attributes, consistently outperforming the second-ranked tracker by at least 2\% in these categories. In the remaining two attributes, ProContEXT slightly surpassed H-DCPT to secure the top position. The distribution of second-place rankings across different attributes shows that HIPTrack led in five attributes, while H-DCPT, ARTrack, and DropTrack each secured second place in two attributes, and ProContEXT in one. By observing the trackers with the lowest performance in these attribute plots, we find that ECO\_HC ranked last in four attributes, AutoTrack and CSWinTT each rank last in three attributes, and SRDCF and ARCF each occupy the lowest position in one attribute. Specifically, H-DCPT achieves scores of 0.732, 0.742, 0.735, 0.784, and 0.748 on the IV, SV, MB, LR, and SOB attributes, respectively, surpassing the second-best baseline, HIPTrack (0.704, 0.714, 0.715, 0.747, and 0.727), by 2.8\%, 2.8\%, 2.0\%, 3.7\%, and 2.1\%, respectively. For the FOC and ARC attributes, H-DCPT scores 0.679 and 0.704, outperforming the runner-up DropTrack (0.658 and 0.663) by 2.1\% in both cases. Regarding the VC and LAI attributes, H-DCPT's scores of 0.731 and 0.725 exceeds those of the second-placed DropTrack (0.707 for both) by 2.4\% and 2.0\%. In the POC attribute, H-DCPT’s score of 0.740 surpasses the second-best performer, ProContEXT (0.713), by 2.7\%. Notably, while ProContEXT slightly outperforms H-DCPT in the OV and ROT attributes with scores of 0.738 and 0.678, the differences are minimal, just 0.01\% and 0.07\%, respectively. These results validate the effectiveness of our proposed approach, which combines darkness clue and historical prompt information to enhance the tracking of low-light objects.

\begin{figure*}[!t]
\centering
\includegraphics[width=1\textwidth]{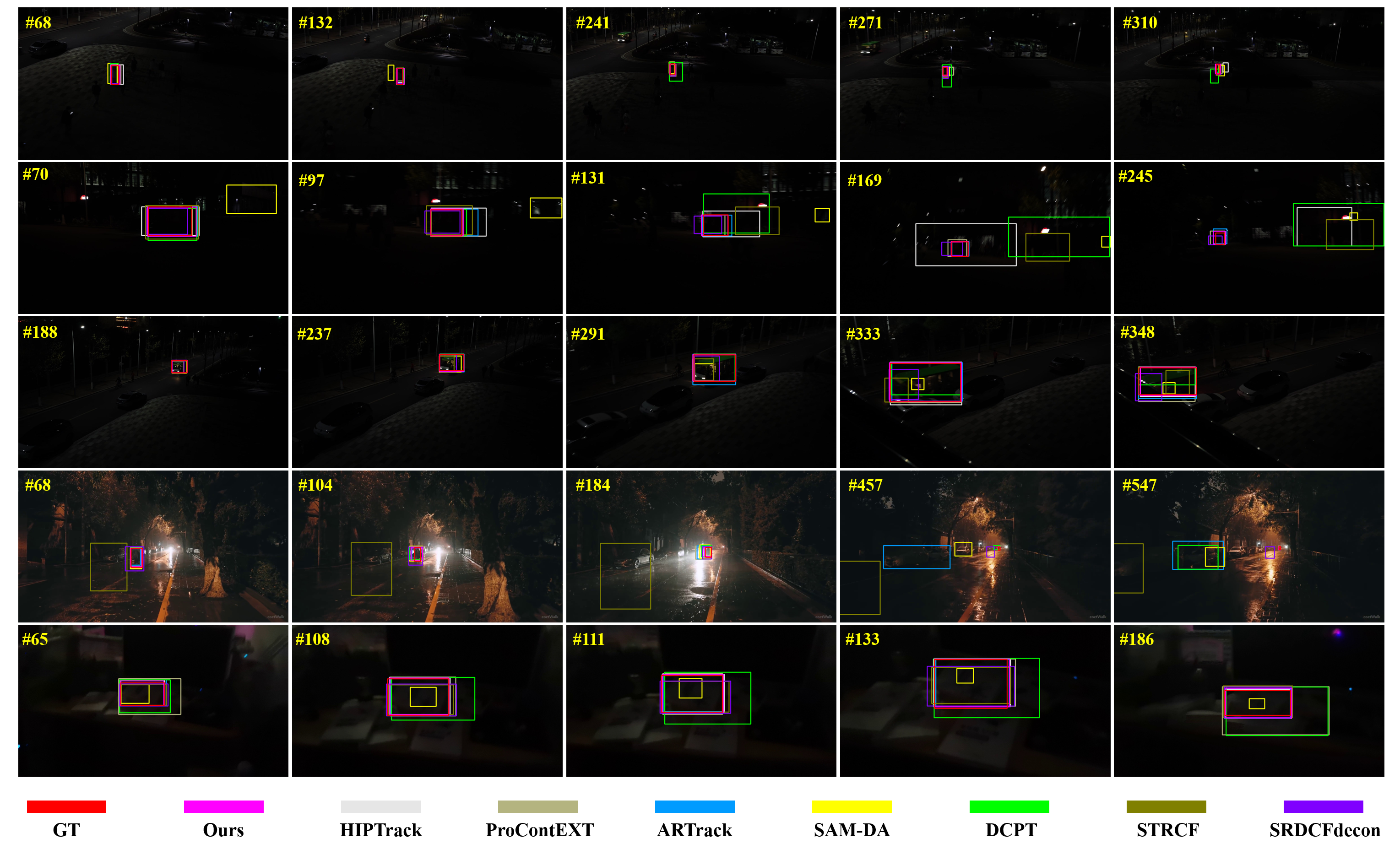}
\caption{Qualitative evaluation on 5 sequences from LLOT, i.e. Walking\_4, Tricycle\_1, Tourbus\_2, Ebikeriding\_3 and Book\_4 from top to bottom. The results of different trackers have been shown with different colors, and 'GT' denotes the ground truth.}
 \vspace{-0.15in}
\label{fig:evaluation}
   
\end{figure*}

\begin{table*}
\centering
\renewcommand{\arraystretch}{1.2} 
\caption{Illustrates the impact of the number of layers in the darkness clue prompt module on H-DCPT's performance on the LLOT dataset, measured by success (S$_{AUC}$), precision (P), and normalized precision (P$_{Norm}$). \textcolor{red}{\textbf{Red}}, \textcolor{blue}{\textbf{blue}}, and \textcolor{green}{\textbf{green}} represent the first, second, and third best performances, respectively.}
\begin{tabular}{c|cccccccccccc} 
\toprule
Layers   & 1    & 2    & 3    & 4    & 5    & 6    & 7    & 8    & 9                               & 10   & 11                                        & 12                              \\ 
\hline
S$_{AUC}$        & 0.559 & 0.565 & 0.563 & 0.565 & 0.570 & 0.564 & 0.571 & 0.566 & \textcolor{blue}{\textbf{0.575}} & 0.572 & \textcolor{green}{\textbf{0.573}} & \textcolor{red}{\textbf{0.576}}  \\
P       & 0.616 & 0.625 & 0.634 & 0.631 & 0.673 & 0.664 & 0.671 & 0.626 & \textbf{\textcolor{blue}{0.676}} & 0.673 & \textcolor{green}{\textbf{0.675}} & \textbf{\textcolor{red}{0.684}}  \\
P$_{Norm}$  & 0.712 & 0.720 & 0.722 & 0.718 & 0.727 & 0.714 & 0.727 & 0.721 & \textbf{\textcolor{blue}{0.732}} & 0.728 & \textcolor{green}{\textbf{0.729}} & \textcolor{red}{\textbf{0.739}}  \\
\bottomrule
\end{tabular}
\label{tab:layer}
\vspace{-0.12in}
\end{table*}

\noindent\textbf{Qualitative evaluation:}  Fig. \ref{fig:evaluation} presents a qualitative comparison of tracking results between our proposed method and seven top-performing trackers. These comparative trackers include: the three best-performing trackers from Table \ref{tab:overallperformance} (HIPTrack \cite{cai2024hiptrack}, ProContEXT \cite{lan2023procontext}, and ARTrack \cite{wei2023artrack}), two algorithms specifically designed for nighttime tracking (SAM-DA \cite{fu2023samda} and DCPT \cite{zhu2023dcpt}), and two of the best-performing DCF-based trackers (STRCF \cite{li2018strcf} and SRDCFdecon \cite{2016srdcfdecon}). The results demonstrate that our proposed H-DCPT maintains robust performance even when faced with challenges such as Illumination Variation (IV), Scale Variation (SV), Motion Blur (MB), Low Resolution (LR), Partial Occlusion (POC), Aspect Ratio Change (ARC), Similar Object(SOB), and Low Ambient Intensity (LAI). Specifically, in the Walking\_4 sequence, only H-DCPT, ARTrack, STRCF, and SRDCFdecon successfully track the pedestrian target. In the Tricycle\_1, Tourbus\_2, and Book\_4 sequences, while many trackers are affected by factors such as illumination variation or motion blur, resulting in inaccurate tracking, H-DCPT consistently demonstrates visually superior results. In the Ebikeriding\_3 sequence, only H-DCPT, HIPTrack, and ProContEXT are able to correctly label the target, while other trackers fail to accurately track it. These qualitative comparisons strongly support the efficacy and superiority of our proposed approach, which combines darkness clue prompts with historical prompt information for tracking objects in low-light conditions. The results demonstrate the significant performance advantages of our method in challenging low-illumination environments.

\begin{figure}[!t]
\centering
\includegraphics[width=3.5in]{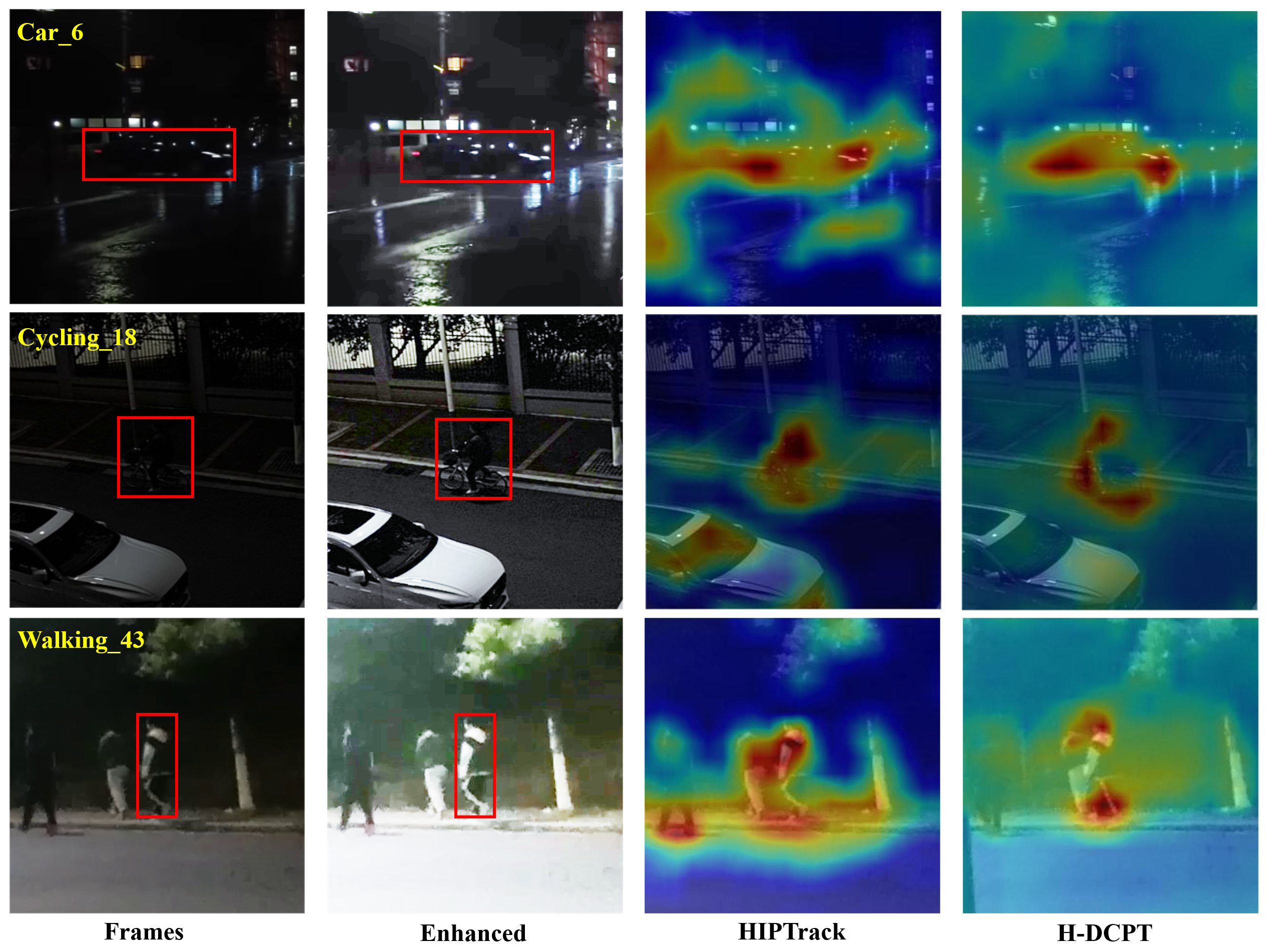}
\caption{Visual comparison of confidence maps generated by the baseline \cite{cai2024hiptrack} and the proposed H-DCPT. The target objects are marked with red bounding boxes. The second column shows images enhanced by the SCI \cite{ma2022sci}. In low-light environments, H-DCPT demonstrates more accurate perception capabilities compared to the baseline.}
\label{fig:attention}
\end{figure}

\subsection{Ablation Study}
\noindent\textbf{Visualization. } As illustrated in Fig. \ref{fig:attention}, we visualize confidence maps from H-DCPT and its baseline, HIPTrack \cite{cai2024hiptrack}. In low-light environments, the baseline model struggles to focus effectively on objects under unfavorable lighting conditions. In contrast, H-DCPT significantly enhances the baseline's nighttime perception capabilities, resulting in satisfactory tracking performance in low-light scenarios. The superior performance of H-DCPT can be attributed to its innovative integration of darkness cue prompts and historical prompt information, which allows for more robust feature representation in low-light scenarios. These results underscore the effectiveness of our proposed method in addressing the challenges of low-light object tracking.

\noindent\textbf{Impact of the number of layers in the darkness clue prompt module:} To investigate the impact of the number of layers in the darkness clue prompt module on H-DCPT's performance on the LLOT dataset, we incrementally increase the number of layers in the darkness clue prompt module from 1 to 12 and observe its effect on the tracker's performance. The experimental results are presented in Table \ref{tab:layer}. As evident from the table, with the increase in the number of layers, the contribution of the darkness clue prompt module gradually enhanced, leading to an overall upward trend in H-DCPT's S$_{AUC}$, P, and P$_{Norm}$ metrics. This finding suggests that increasing the number of layers in the darkness clue prompt module can effectively improve H-DCPT's tracking performance under low-light conditions, demonstrating the effectiveness of combining darkness clue prompts with historical prompt information. However, we also observe that the improvement in S$_{AUC}$ tends to plateau after reaching a certain number of layers. Notably, when the number of layers is 12, H-DCPT achieves its best performance with S$_{AUC}$, P, and P$_{Norm}$ values of 0.576, 0.684, and 0.739, respectively. Based on these results, we adopt 12 layers as the default configuration for H-DCPT.

\section{Conclusion}

In this study, we introduce LLOT (Low-Light Object Tracking), a high-quality benchmark for single object tracking in low-light conditions. LLOT comprises 269 indoor and outdoor low-illumination video sequences, totaling over 132,000 frames. To our knowledge, this is the first ground-view benchmark specifically designed for low-light object tracking. To evaluate the performance of existing trackers and establish a baseline for future comparisons, we conducted a comprehensive evaluation of 39 state-of-the-art tracking algorithms, including both deep learning-based and DCF-based methods. Extensive experiments on LLOT reveal that there is still significant room for improvement in the field of low-light tracking. Furthermore, we propose a novel tracker, H-DCPT, which innovatively combines darkness clue prompt and historical prompt information to achieve superior feature representation. Experimental results demonstrate that H-DCPT significantly outperforms existing state-of-the-art algorithms under low-light conditions.

\vfill
\bibliographystyle{IEEEtran}
\bibliography{main.bib}
\end{document}